\pdfoutput=1

\documentclass[11pt]{article}

\usepackage[final]{acl}

\usepackage{times}
\usepackage{latexsym}

\usepackage[T1]{fontenc}

\usepackage[utf8]{inputenc}

\usepackage{microtype}

\usepackage{inconsolata}

\usepackage{graphicx}
\usepackage{subfigure}

\usepackage{hyperref}

\usepackage{algorithm}
\usepackage[endLComment=]{algpseudocodex}

\usepackage{amsmath}
\usepackage{amssymb}
\usepackage{mathtools}
\usepackage{amsthm}
\usepackage{bm}
\usepackage{afterpage}
\usepackage{amsfonts}
\usepackage{multirow}
\usepackage{cleveref}
\crefformat{section}{\S#2#1#3}
\crefformat{subsection}{\S#2#1#3}
\crefformat{subsubsection}{\S#2#1#3}
\date{\vspace{-5ex}} 
\usepackage{pifont}
\usepackage{latexsym}
\usepackage{graphicx}
\usepackage{color}
\usepackage{subfigure}
\usepackage{multicol}
\usepackage{multirow}
\usepackage{booktabs}
\usepackage{dsfont}

\usepackage{xspace}
\usepackage{xcolor} 
\usepackage[normalem]{ulem}

\usepackage{xspace}
\makeatletter
\newcommand{\smartperiod}{\@ifnextchar.{}{.\@\xspace}}
\newcommand{\smartcomma}{\@ifnextchar.{}{,}\xspace}
\makeatother
\newcommand{\latin}[1]{#1}  
\newcommand{\eg}{\latin{e.g.}\smartcomma}
\newcommand{\ie}{\latin{i.e.}\smartcomma}

\newcommand{\probP}{\mathbb{P}}
\newcommand{\wav}{\textsc{Wav2Vec2.0}}
\newcommand{\modelname}{STAR}
\newcommand{\readaction}{\textsc{read}\xspace}
\newcommand{\writeaction}{\textsc{yield}\xspace}
\newcommand{\fire}{\textsc{fire}\xspace}
\newcommand{\waitk}{\textsc{wait}-$k$\xspace}

\newcommand{\librispeech}{\text{LibriSpeech}\xspace}
\newcommand{\libritts}{\text{LibriTTS}\xspace}

\usepackage[]{cleveref}
\crefname{footnote}{footnote}{footnotes}   
\crefname{equation}{equation}{equations}   
\crefname{corollary}{Corollary}{Corollaries}  
\crefname{line}{line}{lines}               
\crefrangeformat{line}{lines #3#1#4--#5#2#6}
\crefname{lstlsting}{Listing}{Listings}   
\crefname{section}{\S}{\S\S}
\Crefname{section}{\S}{\S\S}    
\crefformat{section}{#2\S#1#3}  
\Crefformat{section}{#2\S#1#3}
\crefrangeformat{section}{\S\S#3#1#4--#5#2#6}
\Crefrangeformat{section}{\S\S#3#1#4--#5#2#6}
\crefmultiformat{section}{\S#2#1#3}{ and~\S#2#1#3}{, \S#2#1#3}{ and~\S#2#1#3}
\Crefmultiformat{section}{\S#2#1#3}{ and~\S#2#1#3}{, \S#2#1#3}{ and~\S#2#1#3}
\crefrangemultiformat{section}{\S\S#3#1#4--#5#2#6}{ and~\S\S#3#1#4--#5#2#6}{, \S\S#3#1#4--#5#2#6}{ and~\S\S#3#1#4--#5#2#6}
\Crefrangemultiformat{section}{\S\S#3#1#4--#5#2#6}{ and~\S\S#3#1#4--#5#2#6}{, \S\S#3#1#4--#5#2#6}{ and~\S\S#3#1#4--#5#2#6}
\theoremstyle{plain}

\theoremstyle{definition}

\theoremstyle{remark}

\usepackage{enumitem}

\definecolor{myDeepYellow}{rgb}{0.9412, 0.6902, 0.302}
\definecolor{myYellow}{rgb}{0.9765, 0.8824, 0.7255}
\definecolor{myBlue}{rgb}{0.6353, 0.7686, 0.8627}

\newcommand\worse{\def\argi{}\docommandworse}
\def\docommandworse#1 {\colorbox{myYellow!80}{#1} \let\next\argi}
\def\argi{\let\next\docommandworse}

\newcommand\worst{\def\argii{}\docommandworst}
\def\docommandworst#1 {\colorbox{myDeepYellow!70}{#1} \let\next\argii}
\def\argii{\let\next\docommandworst}

\newcommand\better{\def\argii{}\docommandbetter}
\def\docommandbetter#1 {\colorbox{myBlue!80}{#1} \let\next\argii}
\def\argii{\let\next\docommandbetter}

\def\docommandbest#1 {\colorbox{myBlue}{#1} \let\next\argii}
\def\argii{\let\next\docommandbest}

\title{Streaming Sequence Transduction through Dynamic Compression}

\author{
  Weiting Tan$^{\spadesuit}$ \quad Yunmo Chen$^{\spadesuit}$\quad Tongfei Chen$^{\heartsuit}$ \\
  \textbf{Guanghui Qin}$^{\spadesuit}$\quad \textbf{Haoran Xu}$^{\spadesuit}$\quad \textbf{Heidi C. Zhang}$^{\clubsuit}$ \\
  \textbf{Benjamin Van Durme}$^{\spadesuit}$\quad \textbf{Philipp Koehn}$^{\spadesuit}$ \\
  $^{\spadesuit}$Johns Hopkins University \quad
  $^{\heartsuit}$Microsoft \quad
  $^{\clubsuit}$Stanford University
}

\begin{document}
\maketitle

\begin{abstract}
We introduce STAR (Stream Transduction with Anchor Representations), a novel Transformer-based model designed for efficient sequence-to-sequence transduction over \emph{streams}. STAR dynamically segments input streams to create compressed \emph{anchor} representations, achieving nearly lossless compression (12$\times$) in Automatic Speech Recognition (ASR) and outperforming existing methods. Moreover, STAR demonstrates superior segmentation and latency-quality trade-offs in simultaneous speech-to-text tasks, optimizing latency, memory footprint, and quality.\footnote{~Codes available at: \url{https://github.com/steventan0110/STAR}}
\end{abstract}

\section{Introduction}\label{sec::intro}

Sequence transduction, also referred to as sequence-to-sequence modeling, has shown remarkable success across various domains, including speech translation \cite{Liu2019EndtoEndST, di-gangi-etal-2019-must, Li2020MultilingualST} and automatic speech recognition \cite{prabhavalkar2023endtoend,Li2021RecentAI,conformer}. Traditionally, these models operate under the assumption of fully observing input sequences before generating outputs.
However, this requirement becomes impractical in applications necessitating low latency or real-time output generation such as simultaneous translation \citep[\textit{inter alia}]{ma-etal-2019-stacl, Chang_2022,communication2023seamless}. The concept of streaming sequence transduction \cite{Inaguma2020MinimumLT, Kameoka2021FastS2SVCSN,Chen2021DirectSS,Wang2022LAMASSUAS,Chen2021DirectSS,Xue2022LargeScaleSE}, or stream transduction, arises to address this challenge. Unlike traditional sequence transduction, stream transduction operates on partially observed input sequences while simultaneously generating outputs. This requires deciding when to initiate output generation, a task inherently tied to identifying critical \textit{triggers} within the input sequence.
Triggers mark moments when sufficient input information has been received to initiate output generation, thus minimizing latency. Consequently, they partition the input sequence into discrete \textit{segments}, with outputs accessing only information preceding each trigger.

\begin{figure}[t]
    \centering
    \includegraphics[width=0.9\linewidth]{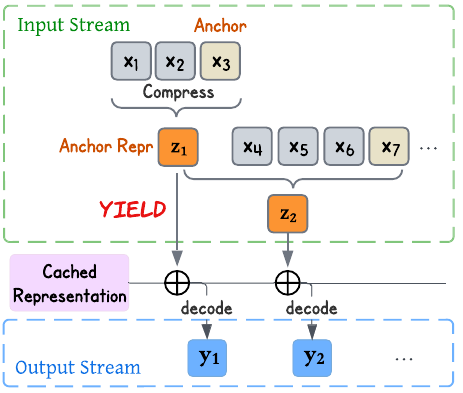}
    \vspace{-1em}
    \caption{When \writeaction is triggered, the current segment's information is compressed into an anchor representation to generate the next output.}
    \vspace{-1.5em}
    \label{figure::teaser}
\end{figure}
Locating these triggers poses a significant challenge. Prior approaches have explored methods that employ fixed sliding windows to determine triggers \cite{ma-etal-2019-stacl,ma-etal-2020-simulmt}, or learning models to predict triggers \cite{Ma2020Monotonic,Chang_2022}, yet timing remains a complex issue. Beyond reducing latency, another challenge for stream transduction is how to efficiently represent historical information while optimizing memory usage. Prior work \citep[\textit{inter alia}]{compressive2020, efficient2022,unlimiformer23} has mostly focused on improving the efficiency of Transformer but does not investigate streaming scenarios. Reducing the memory footprint for streaming systems introduces additional complexity as models must determine when certain information becomes less relevant for future predictions.

In this work, we propose \textbf{S}tream \textbf{T}ransduction with \textbf{A}nchor \textbf{R}epresentations (STAR), a novel approach designed to maximize the benefits of stream transduction, optimizing both generation latency and memory footprint. STAR dynamically segments the input stream into buffers that contain similar levels of information. Then, it introduces the concept of \textbf{anchors}, which aggregate a buffer of information (multiple vector representations) into single-vector anchor representations. Once an anchor representation is yielded, it triggers the generation process to yield another token.

We present a learning strategy to train STAR end-to-end so that the model learns to dynamically select anchor positions with the following objectives: (1) anchor positions are selected such that each segment contains the right amount of information for generating the next output; (2) anchor representation effectively compress the information of its preceding segment. For example, in \cref{figure::teaser}, the model triggers \writeaction at index 3 (which makes it an anchor position), compressing the information of the current chunk $\bm X=(x_1, x_2, x_3)$ into anchor representation $z_1$ to generate output $y_1$. Such a process repeats each time \writeaction is triggered.
To summarize, our contributions are as follows: (1) we propose STAR that dynamically segments and compresses input streams, trading-off among latency, memory footprint, and performance for stream transduction; (2) we validate the effectiveness of our approach on well-established \emph{speech-to-text} tasks. Our results show that STAR greatly outperforms existing methods, obtaining better compression ability and excelling in quality-latency trade-offs.

\section{Methodology}\label{sec::method}
\subsection{Problem Formulation}

In sequence-to-sequence transduction, feature $\bm X =  (\bm x_1, \dots, \bm x_{T_x})$ is normally first extracted from the raw input sequence. Then the decoder can encode and use such features to generate an output sequence $\bm Y = (y_1, \dots, y_{T_y})$. The encoder and decoder can be implemented using various models such as Recurrent Neural Networks \cite{HochSchm97, chung2014empirical,rnn_survey} and Transformers \cite{transformer}, depending on the input and output characteristics. In the context of streaming sequence transduction, where the input (and their features $\bm X$) is partially observed, \textit{a causal encoder and decoder are necessary}. The causal encoder processes the partially observed feature $\bm X_{<\tau}$ ($\tau \le T_x$) to produce their encoding. Suppose the first $k$ outputs are already generated, the causal decoder sample the next output $y_{k+1}$ with $\probP(y_{k+1} | \bm X_{<\tau}, \bm Y_{<k+1}; \theta)$, where $\theta$ represents the parameter set. 

\begin{algorithm}[t]
\small
\caption{High-level overview of STAR}
\begin{algorithmic}[1]
\State \textbf{Input}: Input stream $\bm X$, threshold $\beta$
\State \textbf{Output}: Output stream $\bm Y$
\State \textbf{Initialize}: cached repr. $\bm Z \leftarrow \varnothing$; buffer $\bm B \leftarrow \varnothing$

\While{$y \neq \textsc{eos}$}
\State $\alpha \leftarrow 0; \bm B \leftarrow \varnothing$ \Comment{clear buffer}
\While{$\bm x \leftarrow \textsc{read}(\bm X)$}
\Comment{\readaction new inputs}
\State \textsc{append}($\bm B, \bm x$) \Comment{add to buffer}
\State $\alpha \leftarrow \alpha + F_{\rm seg}(\bm x)$
\If{$\alpha \geq \beta$}\Comment{yield triggered}
\State $\bm H = F_{\rm enc}(\bm B \mid \bm Z)$ \Comment{encode segment buffer}
\State $\bm z = \textsc{compress}(\bm H)$
\State \textsc{append}($\bm Z, \bm z)$ \Comment{embedding for segment}
\State $y \leftarrow F_{\rm dec}(\cdot | \bm Y, \bm Z)$
\State \textbf{yield} $y$
\State \textbf{break}
\EndIf
\EndWhile
\EndWhile
\end{algorithmic}
\label{alg::overview}
\end{algorithm}

Deciding when to generate (yield) a new token is the core of streaming sequence transduction where a segmenter/predictor \cite{streaming_asr, Chang_2022} is typically trained to control timing for yield operation. Our approach to tackling stream transduction is outlined in \cref{alg::overview}. It involves a learnable segmenter that scores the importance of each input feature to decide if enough information has been accumulated in the current buffer of features. As the segmenter scores input feature in a frame-wise fashion (\cref{alg::overview}, line 8), we accumulate the scores $\alpha$ until it reaches a pre-defined threshold $\beta$. When the threshold is reached, it indicates that enough information has been accumulated in the current buffer $\bm B$ of features. Subsequently, we compress the features into a single vector representation $\bm z$ that we call \textbf{anchor representation} (line 11). $\bm z$ is computed for each buffer and cached into the history anchors $\bm Z$, which is then conditioned by the decoder to generate new tokens (lines 12-13). The details of our segmentation and compression mechanism are introduced in \cref{sec::method_compression}.

\subsection{Segmentation with Dynamic Compression}\label{sec::method_compression}
In this section, we provide details of different components in \cref{alg::overview}. We first describe how to learn the segmenter $F_{\rm seg}(\cdot)$ with feedback from the encoder-decoder's cross-attention. Then we present how anchor representations are obtained through our selection-based compression method.

\begin{figure}[t]
    \centering
    \includegraphics[width=0.9\linewidth]{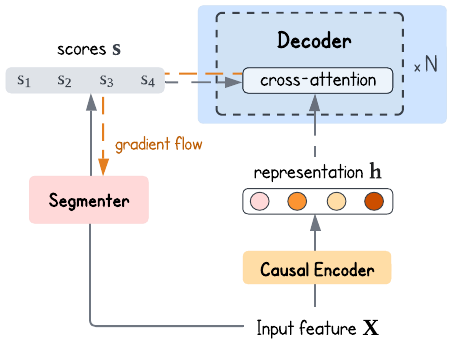}
    \vspace{-0.5em}
    \caption{Visualization for the training of the segmenter through feedback from the encoder-decoder's cross-attention.}
    \label{figure::prediction_network}
    \vspace{-1.2em}
\end{figure}

\paragraph{Learning Segmenter with Cross-attention} 
We propose a learnable segmenter trained with feedback from the encoder-decoder cross-attention. Following \cref{alg::overview}, a segmenter is used to evaluate (score) input features as they are read into the system. Such scores $\bm s = F_{\rm seg}(\bm X)$ are then used to determine if \writeaction is triggered (\ie whether to segment streams). Effective segmentation is crucial in streaming sequence transduction to avoid sub-optimal transformation due to premature triggering or increased latency from delayed output. Since the ideal segmentation depends on several factors (the input's information density, the input and output's modalities, and the task at hand, \emph{etc.,}),  we rely on the cross-attention between the encoder and decoder to guide the segmenter (shown in \cref{figure::prediction_network}). 

Specifically, we follow cross-attention from Transformers \cite{transformer} to use three projections $\bm W_\text{Q}, \bm W_\text{K}, \bm W_\text{V}$ to generate the query vector $\bm y \bm W_\text{Q} \in \mathbb{R}^{T_y \times d}$,  the key vector $\bm h \bm W_\text{K} \in \mathbb{R}^{T_x\times d}$  and the value vector $\bm h \bm W_\text{V} \in \mathbb{R}^{T_x\times d}$ (where $d$ is the dimensionality of the representation) and compute cross-attention as:
\begin{equation}
    \label{eq::cross_attn}
     S(\bm h, \bm y) = (\bm h \bm{W}_\text{k}) (\bm y \bm{W}_\text{Q})^T 
\end{equation}

Then, as illustrated in \cref{figure::prediction_network}, we \textbf{inject} segmenter's scores into it the cross attention:
\begin{equation}
 \tilde S(\bm h, \bm y) =  S(\bm h, \bm y) + F_{\rm seg}(\bm x)
    \label{eq::injection}
\end{equation}
The updated cross-attention $\tilde S(\bm h,\bm y)$ is then used to transform the value vector $\bm W_\text{V}$ and will be used by the decoder to compute the loss function.
Since the segmenter's scores are injected in \cref{eq::injection}, \emph{it can be updated with end-to-end back-propagation}. Specifically, suppose the loss objective $\mathcal{L}$ is computed, with the chain rule, we have the gradient for the predicted score $\bm \alpha = F_{\text{seg}}(\bm x)$ as:

\begin{equation}
\begin{aligned}
    \frac{\nabla \mathcal{L}}{\nabla \bm{\alpha}} &= \sum_{i=1}^l \frac{\nabla \mathcal{L}}{\nabla \tilde{S}^i(\bm{h}, \bm{y})} \cdot \frac{\nabla \tilde{S}^i(\bm{h}, \bm{y})}{\nabla \bm{\alpha}}\\
    &= \sum_{i=1}^l \frac{\nabla \mathcal{L}}{\nabla \tilde{S}^i(\bm{h}, \bm{y})} \cdot  \frac{\nabla}{\nabla \bm{\alpha}} \left( S(\bm h, \bm y) + \bm{\alpha} \right) \\
    &= \sum_{i=1}^l \frac{\nabla \mathcal{L}}{\nabla \tilde{S}^i(\bm{h}, \bm{y})} \nonumber
\end{aligned}
\end{equation}

where $l$ is the number of transformer layer and $\tilde{S}^i(\bm h, \bm y)$ is the cross-attention for $i^{\text{th}}$ layer. We observe that the gradient impacting the segmenters is directly proportional to the gradient on the cross-attention logits. Consequently, by injecting cross-attention, we can train segmenters to prioritize positions that are more significant to the decoder. 

After training the segmenter, we predict scores $\bm s = F_{\rm seg}(\bm x)$ for input features and use the scores to segment the input sequence. Note that the predicted scores can be used differently based on the task. In the special case where the whole sequence is fully observed (\ie regular non-streaming tasks), we do not \writeaction output anymore. Instead, we simply select the top $k$ scoring positions as anchors and use their representation for the decoder to generate outputs, as formalized below ($I$ is a set of indices):
\begin{align}
    I &= \textsc{SelectTop}_k(\bm s) \label{eq::topk_indice}\\
    \bm H &= F_{\rm enc}(\bm x) \in \mathbb{R}^{T_x \times d}\\
    \bm Z &= \bm H[I] \in \mathbb{R}^{k \times d}\label{eq::topk_select}
\end{align}
The compression rate is then $r = T_x / k \in [1, \infty)$ assuming $k \leq T_x$. In a more general case where streaming is enabled, the score is commonly accumulated \cite{Inaguma2020MinimumLT, Ma2020Monotonic} until a certain threshold is reached. We use a threshold $\beta=1$ throughout experiments. Specifically, we first scale $\bm s$ to $[0,1]$ range values $\bm \alpha = \text{sigmoid}(\bm s)$ and accumulate $\alpha$ following \cref{alg::overview} (line 8) to \writeaction new output. The accumulation of scores is a natural way to ensure a similar level of information is contained in each buffer. This corresponds to a larger buffer when the sound signal is sparse (see \cref{app::segmentation_analysis} for visualization), which gives better latency-quality control.

\paragraph{Compression with Anchor Representation} 

\begin{figure}[t]
    \centering
    \includegraphics[width=0.9\linewidth]{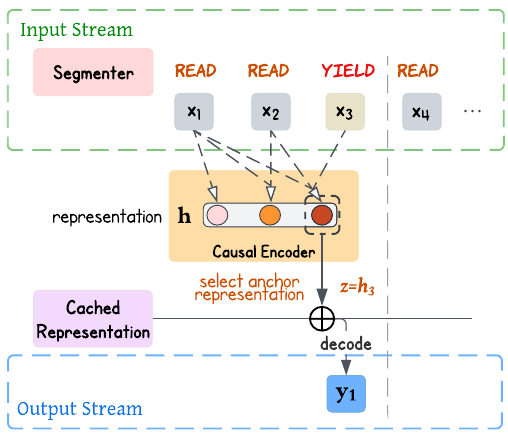}
    \vspace{-0.5em}
    \caption{Visualization for the proposed ``selection as compression'' method. Input features are transformed by the encoder and we only select the encoding at the anchor position (where \writeaction is triggered) as the compressed representation.}
    \label{figure::compressor}
    \vspace{-1.5em}
\end{figure}

Every time an anchor is predicted by our trained segmenter, the model triggers generation with some buffer $\bm B \in \mathbb{R}^{b \times d}$ of b features. Subsequently, we transform such features into a high-dimensional representation $\bm H \in \mathbb{R}^{b \times d}$ with a \textbf{causal} encoder\footnote{~In practice, we are inspired by BERT \cite{Devlin2019BERTPO} to add a special type embedding $\bm e$ to anchor tokens before passing through the encoder}. The causality of such an encoder ensures that representations at later positions contain information only from earlier positions. Then, we \textbf{only select} the representation at the anchor position (the last index of the current buffer) $\bm z = \bm H[b] \in \mathbb{R}^{1\times d}$ to represent the information of the whole buffer $\bm B$. Selected representations are also called anchor representations/vectors. For example, in \cref{figure::compressor}, \writeaction is triggered at index $3$; therefore we first transform the features into representations $\bm H = F_{\rm enc}(\bm B | \bm Z)$, and select $\bm H[3]$ as the anchor vector $\bm z$ to decode the next output with cached representation $\bm Z$.

\subsection{Model Training}
To train models for streaming sequence transduction, we primarily rely on the conventional objective -- negative log-likelihood (NLL) loss:
\begin{equation}
\begin{aligned}
    \mathcal{L}_{\text{NLL}}(\bm X, \bm Y, \theta) &= -\log \probP(\bm Y|\bm X;\theta)\\
    &= -\sum_{t=1}^{T_y} \log \probP(y_t|\bm Y_{<t},\bm Z_{<t}; \theta)
\label{eq:nll}
\end{aligned}
\end{equation}
Note that the loss is defined over $\bm X, \bm Y$ as both input and output sequences are fully observed during training. In addition, the loss defined in \cref{eq:nll} is slightly different than regular NLL in that the decoder can only use representation observed so far ($\bm Z_{<t}$) to generate the $t^{\text{th}}$ output. This method is also referred to as Infinite-Lookback \citep[\text{IL}]{milk,liu-etal-2021-cross} and is used to mitigate the train-test mismatch as future representation cannot be observed during inference. Besides using NLL to update the encoder and decoder, we also follow prior work \cite{Chang_2022} to regularize the segmenter so that the number of \writeaction is the same as the output length $T_y$. Due to page limitations, we refer readers to \cref{app::cif} for more details.

\section{Experiments: Non-Streaming Compression}\label{sec::experiments_no_stream}
We experiment on the non-streaming ASR task to better demonstrate the effectiveness of our selection-based compression method, since we do not need to consider the quality-latency trade-off as in the streaming scenario. We compare our method with other common baselines like Convolutional Neural Networks \citep[CNN]{cnn,conv} and Continuous Integrate and Fire \citep[CIF]{dong2020cif}. 

\paragraph{Datasets and Evaluation Metrics}
We conduct experiments on the LibriSpeech \cite{librisppech_dataset} and LibriTTS \cite{zen2019libritts} dataset's ``Clean-360h'' section, which contains 360 hours of speech and their corresponding transcriptions. To evaluate ASR performance, we compute the word error rate \citep[\text{WER}]{wer} between reference transcriptions and the generated text.

\subsection{Training Setup}
\paragraph{Compression with Anchor Representations}
In \cref{sec::method}, we propose a general approach for stream transduction with dynamic compression. Now we instantiate the framework for the ASR task. We first use \wav~\cite{w2v2} to extract features $\bm X$ from the input speech sequence. We then use a 4-layer decoder-only Transformer\footnote{~Following the implementation of GPT2 from Huggingface \url{https://huggingface.co/gpt2}} as our \textbf{causal encoder} for compression, from which we select out anchor representation $\bm z$. The segmenter is implemented with a 2-layer Feed-Forward Network. For the decoder, we use a 4-layer decoder-only Transformer with an additional linear layer as the language modeling head. For details of hyperparameters, we direct readers to \cref{app::hyper}.

As described in \cref{sec::method}, we train the Encoder-Decoder model with a segmenter learned through cross-attention feedback. Given the extracted feature $\bm X = (x_1, x_2, \cdots, x_{T_x})$ and a target compression rate $r \in [1, \infty)$, we select top $k = T_x / r$ scoring positions and use their encodings as anchor representations (following \cref{eq::topk_select}). We then feed the anchor representation $\bm Z$ to the decoder to generate text tokens. In practice, most input speeches from LibriTTS are less than 10 seconds, corresponding to a feature sequence of length $T_x = 10 * 16000 / 320 = 500$ (with a standard sampling rate 16 kHz and \wav ~has a stack of CNNs that reduce input sequence by 320$\times$). Therefore, we chose some reasonable compression rates (\ie $r = 12, 18, 30$) to test our compression methods. We now briefly describe two baselines that we compared against: CNNs and CIF.

\paragraph{Baseline: CNN} A simple compression component is CNN. After we obtain speech feature $\bm X$, we apply CNNs with pre-defined strides to compress the feature. The encoder (a vanilla Transformer-Encoder module without our selection-based compression) further transforms such compressed features into encoder representations for the decoder to generate outputs. To enhance the capacity of CNNs, we follow \citet{zeghidour2021soundstream, encodec} to add two CNNs with kernel size $(5, 1)$ and stride size $(1, 1)$ as residual connection. More details about CNNs and their configurations are available in \cref{figure::conv_design} (in \cref{app::hyper}).

\paragraph{Baseline: CIF}
Continuous Integrate and Fire \cite{dong2020cif,dong-etal-2022-learning, Chang_2022} uses a neural network to predict scores for each position and accumulates the scores until a threshold is reached, thereafter triggering the generation of a new token (called \fire by the original paper). For each segment, CIF averages representations in the segment by directly weighing them with the predicted scores. For a fair comparison with prior work, we adopt the implementation from \citet{dong-etal-2022-learning} into our codebase.


There are two major differences between our method and CIF: firstly, STAR segmenter leverages cross-attention between encoder-decoder to interactively update representations, whereas CIF employs a weighted average of representations solely from the encoder side; secondly, STAR pushes information to condense in particular anchor at \writeaction positions and performs explicit selections, whereas CIF's representations are averaged across each segment. Broadly, these distinctions mirror the differences between hard and soft attention mechanisms \cite{sft_hard_attn, luong-etal-2015-effective}. We refer readers to \cref{app::cif} and the original paper \cite{dong2020cif} for more details.

\subsection{Results of Different Compression Methods}\label{sec::compress_asr_result}

\begin{figure}[t]
    \centering
    \includegraphics[width=0.96\linewidth]{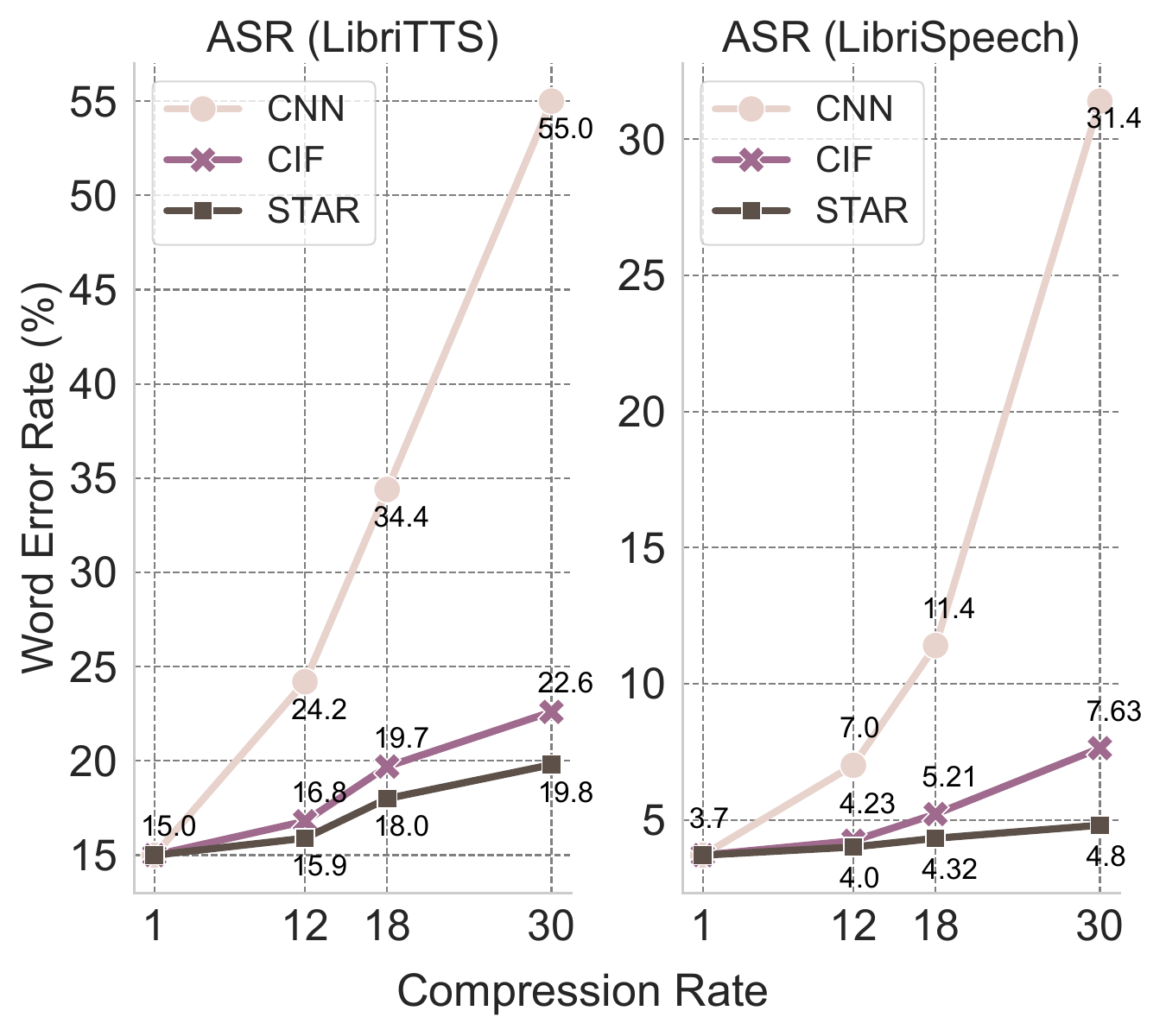}
    \vspace{-1em}
    \caption{ASR performance (evaluated by WER) by different compression methods. From the figure, \modelname ~outperforms other compressors and the gap enlarges as the compression rate increases.}
    \label{figure::compression_asr}
    \vspace{-1em}
\end{figure}

We test the compression performance on three compression rates $r \in \{12, 18, 30\}$. As shown in \cref{figure::compression_asr}, our compression module obtains the best performance, achieving almost lossless compression when $r=12$, and consistently outperforms the other two methods on different compression rates. By comparing the trend in detail, we find that CNNs are sub-optimal as the compressor because they operate on a small local window and change the underlying feature representation, which might be hard for the encoder and decoder to adapt to. Now comparing CIF and \modelname. As the compression rate increases, the gap between \modelname ~and CIF also increases. When $r=30$, \modelname ~outperforms CIF by about 3 WER points on both \librispeech and \libritts. From the results, we have verified that \modelname ~is more effective in compressing representation compared to CNN and CIF. Later in our analysis (see \cref{sec::analysis}), we provide evidence of \modelname ~achieving more robust compressed representations. Lastly, to exclude the influence from the text decoder, we also designed a speech similarity task in \cref{app::sts} to show that \modelname ~results in better-compressed speech representation.

\begin{figure*}[t]
    \centering
    \includegraphics[width=0.98\linewidth]{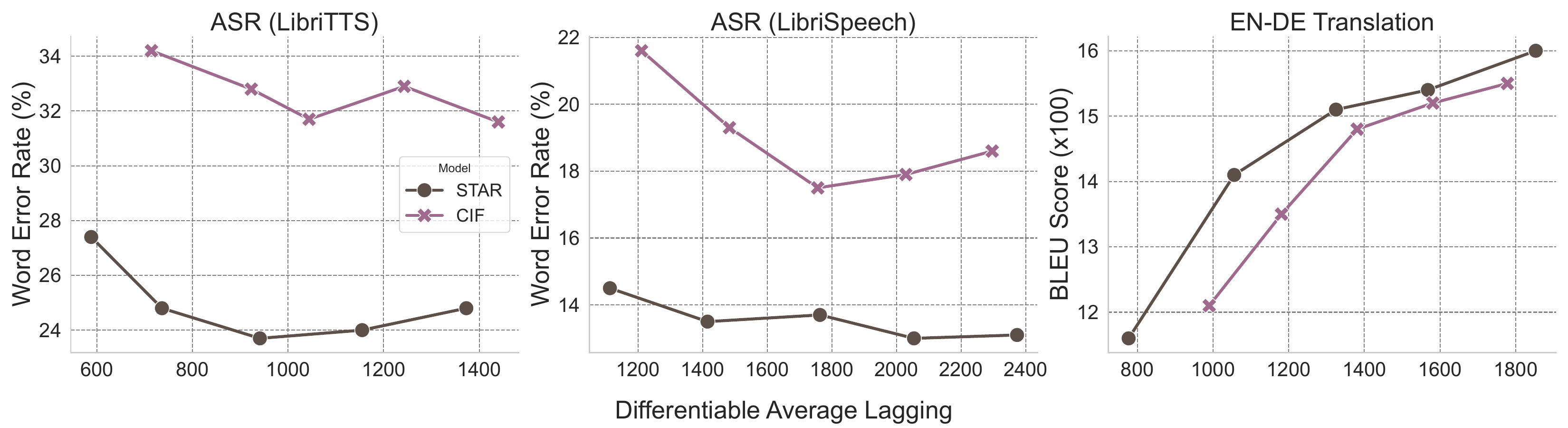}
    \vspace{-0.5em}
    \caption{Lateny-Quality trade-off for CIF and \modelname. The five markers on the line correspond to different \waitk strategies (from left to right, \waitk $\in \{1,2,3,4,5\}$).}
    \label{figure::simul_flexible_asr}
    \vspace{-1em}
\end{figure*}

\section{Streaming Experiments: Simultaneous Speech Recognition and Translation}\label{sec::experiments_stream}

\paragraph{Datasets} For our simultaneous S2T experiments, we use the English-German (EN-DE) portion of the MuST-C V1 \cite{di-gangi-etal-2019-must} dataset for speech translation (ST). We also include results for simultaneous ASR using LibriSpeech and LibriTTS. Note that since our method is based on a general Encoder-Decoder Transformer, it is not tailored to ASR by leveraging monotonic alignment or using small character-level vocabulary.

\paragraph{Evaluation Metric} To evaluate the quality of generated output, we use WER for the ASR task and BLEU \cite{papineni-etal-2002-bleu} for the speech translation task. For simultaneous S2T, latency measurement is essential and we resort to the commonly used metric, Differentiable Average Lagging \citep[\text{DAL}]{milk}, which was originally proposed for simultaneous text translation and later adapted to speech translation in \cite{ma-etal-2020-simuleval}. The smaller the DAL, the better the system in terms of latency. We refer readers to \cref{app::dal} for details on the latency metric.

\paragraph{Experiment Setup}
Our first step is to train an \emph{speech-to-text} (S2T) streaming model without a segmenter. To make \wav ~causal, we add a causal mask and train it jointly with the encoder and decoder until convergence. Once the vanilla streaming S2T model is trained, we freeze the \textbf{causal} \wav ~model as the feature extractor and start fine-tuning the encoder and the decoder with the segmenter.

\begin{figure}[t]
    \centering
    \includegraphics[width=0.9\linewidth]{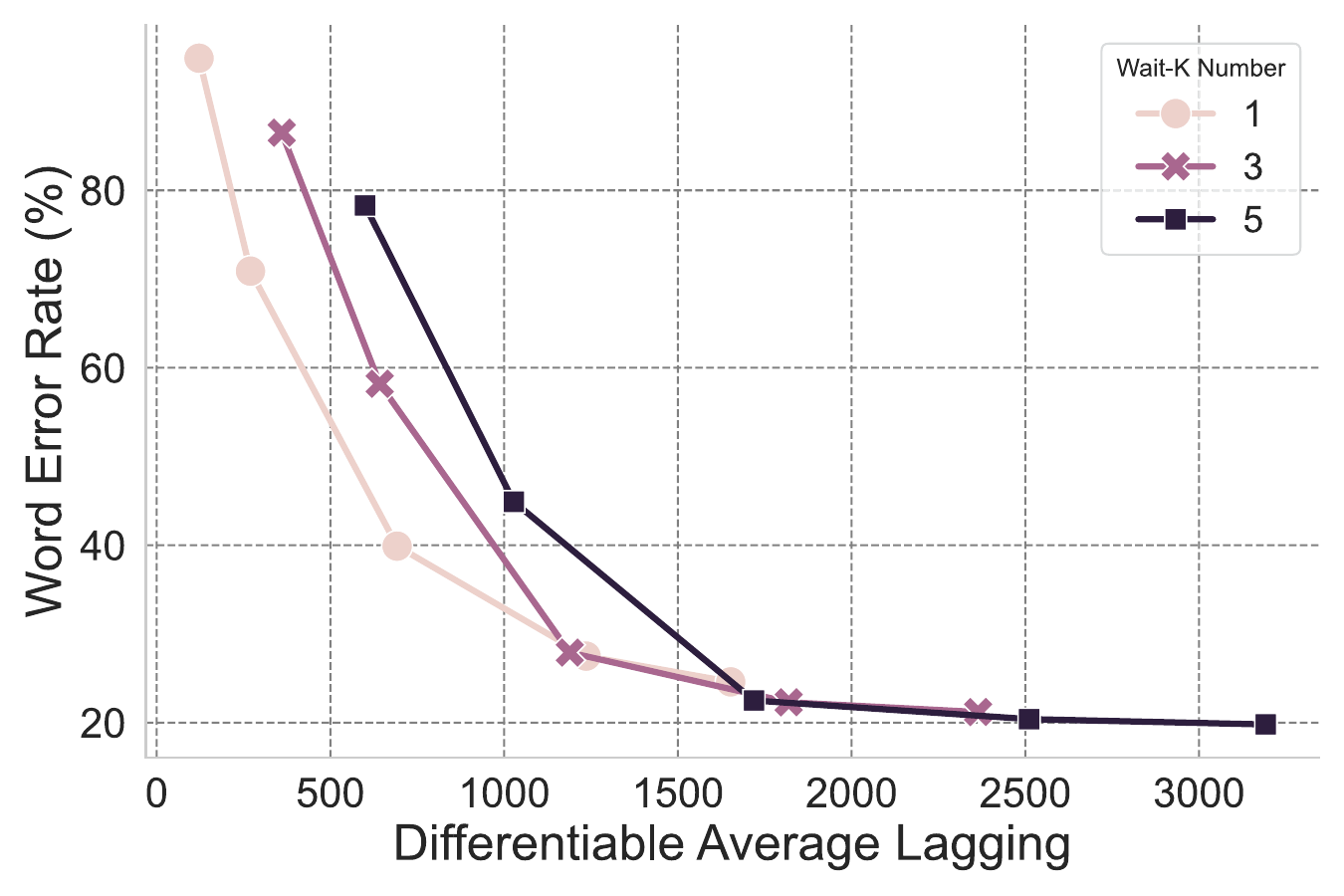}
    \vspace{-0.5em}
    \caption{Quality-latency trade-off for fixed-decision S2T model. Each line corresponds to a different \waitk strategy and each marker corresponds to a stride size of $\{120, 200, 280, 360, 440\}$ms.}
    \label{figure::simul_fix_asr}
    \vspace{-1em}
\end{figure}

\paragraph{Experimental Results}
We show the experiment results in \cref{figure::simul_flexible_asr} where we plot WER/BLEU v.s. DAL to demonstrate the quality-latency trade-off for each system. In our evaluation, we adapt the \waitk policy \cite{ma-etal-2019-stacl} for all systems. Here \waitk denotes the number of speech segments we encode first before decoding text tokens. A larger \waitk value generally results in higher latency but better S2T performance. In our work, we focus on low-latency scenarios where flexible decision policies like CIF and \modelname ~are most useful; Therefore, we set \waitk value to 1 to 5.

We first present the baseline system for simultaneous ASR with a fixed decision policy in \cref{figure::simul_fix_asr}. We use the vanilla streaming S2T model (no compression) and apply a fixed stride size to slide through the speech and generate text tokens. As shown in \cref{figure::simul_fix_asr}, using a large stride like 360ms (\ie each chunk corresponds to a speech feature of length $0.36 \times 16000 /320 = 18$) or 440ms, simultaneous ASR achieved $< 20$ WER. However, the latency is also extremely high (over 2000 DAL). For smaller strides, quality of generated output is suboptimal because not enough information is provided for the text decoder to generate each new token. A flexible decision policy could alleviate such issues and provide better latency-quality trade-off. From \cref{figure::simul_flexible_asr}, we see that for both CIF and \modelname, their output has better quality when the latency is low. For instance, on \libritts, \modelname ~achieves about 24 WER with a DAL smaller than 800 while the best-performing fixed decision policy only obtains such performance with a DAL of about 1200. 

Comparing CIF with \modelname ~across three datasets (\librispeech, \libritts, and MUST-C), we find that \modelname ~consistently achieves better performance, obtaining a lower WER (or higher BLEU) score with relatively lower latency across different \waitk strategies. This demonstrates that \modelname ~gives a better flexible policy to \writeaction new tokens, and the compressed representation encodes more information for target text generation. In \cref{app::segmentation_analysis}, we compare qualitative examples and visualize the difference in the segmentation from CIF and \modelname. Overall, we find the segmentation from \modelname ~better corresponds to the target texts, achieving superior simultaneous S2T performance.

\section{Analysis}\label{sec::analysis}

\subsection{Memory Efficiency}\label{sec::memory}
\begin{figure}[t]
    \centering
    \includegraphics[width=0.9\linewidth]{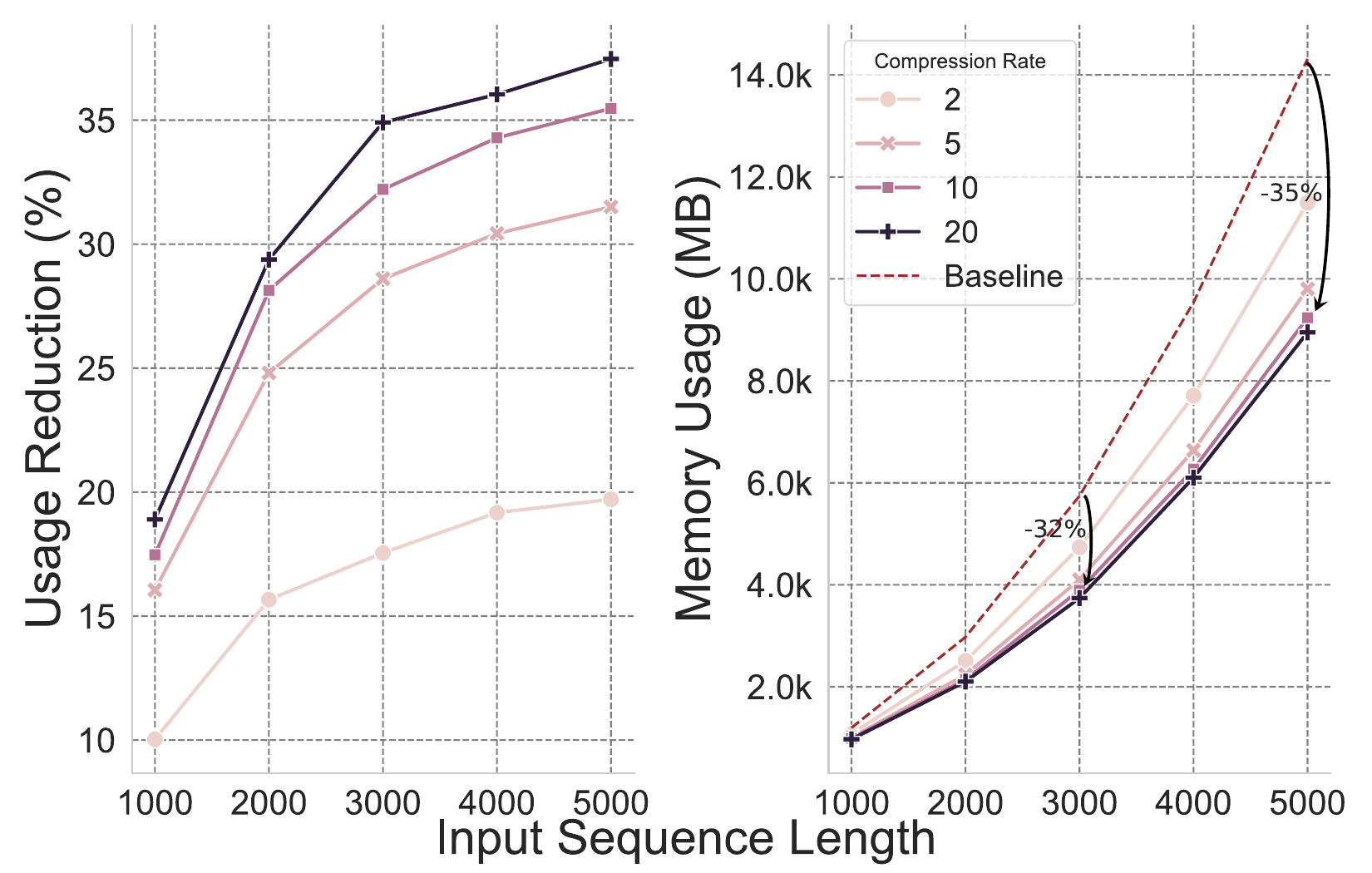}
    \vspace{-0.5em}
    \caption{Memory usage and reduction from our proposed method (with compression rates $r \in \{2,5,10,20\}$). More results and detailed setup are provided in \cref{app::memory}.}
    \label{figure::memory}
    \vspace{-1.5em}
\end{figure}

Since STAR condenses information in each buffer into anchor representation, it enhances memory efficiency by caching compressed representation for the decoder to generate outputs. With a compression rate $r$, a batch size $b$, and input features of average length $T_x$, and hidden dimension $d$, our system compresses the encoder representation from $bdT_x$ to $bdT_x/r$. Besides memory consumption, note that cross-attention computation (\cref{eq::cross_attn}) is quadratic w.r.t. encoder representation's length; thus, our method reduces the cost of its computation by a factor of $r^2$. Besides theoretical analysis, we benchmark the actual memory usage and the percentage of usage reduction achieved by different compression rates. From \cref{figure::memory},  we show that with a rate of $r=10$ (which achieves nearly lossless compression), STAR reduces the memory consumption by more than 30\% when transducing an input feature of length longer than 3,000. For the full details of our benchmark setup and results, we refer readers to \cref{app::memory}.

\subsection{Robustness}\label{sec::robustness}
In this section, we evaluate the robustness of streaming models (CIF and STAR) by subjecting them to compression and segmentation conditions different from their training setup. We find that STAR is more robust than CIF, retaining better transduction when operating on context windows not exposed to during training.

\begin{figure}[t]
    \centering
    \includegraphics[width=0.9\linewidth]{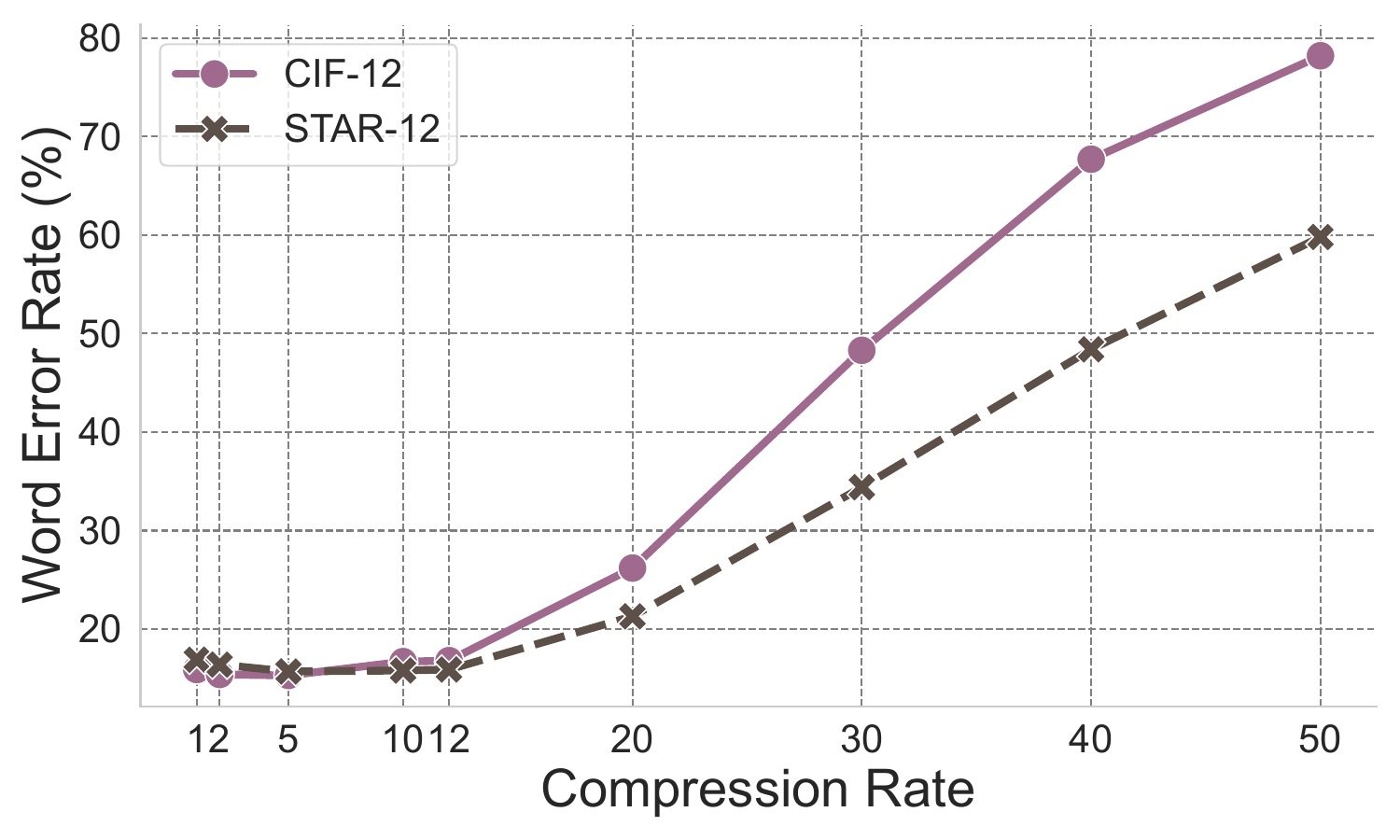}
    \vspace{-1em}
    \caption{CIF and \modelname ~based model trained with compression rate 12 are evaluated on various compression rates (ranges from 1 to 50). For a lower compression rate ($\leq 12$), both models preserve their quality well. For a higher compression rate ($>12$), \modelname ~is more robust and its performance degrades slower than CIF.}.
    \label{figure::diff_compression}
    \vspace{-2em}
\end{figure}

\paragraph{Various Compression Rates at Inference}

As detailed in \cref{sec::experiments_no_stream}, we trained CIF- and \modelname-based models with a compression rate of $r=12$ (denoted as CIF-12 and \modelname-12) and tested them under varying compression rates. Both models perform well at $r \leq 12$, as expected since they are trained for 12$\times$ compression. However, when $r > 12$, \modelname-12 shows significantly less degradation compared to CIF-12, indicating superior retention of information. This resilience arises from \modelname’s design, which focuses information into anchor positions, ensuring each anchor retains substantial information even at higher compression rates. In contrast, CIF’s averaging approach leads to increased interference between representations.

\begin{figure}[t]
    \centering
    \includegraphics[width=0.9\linewidth]{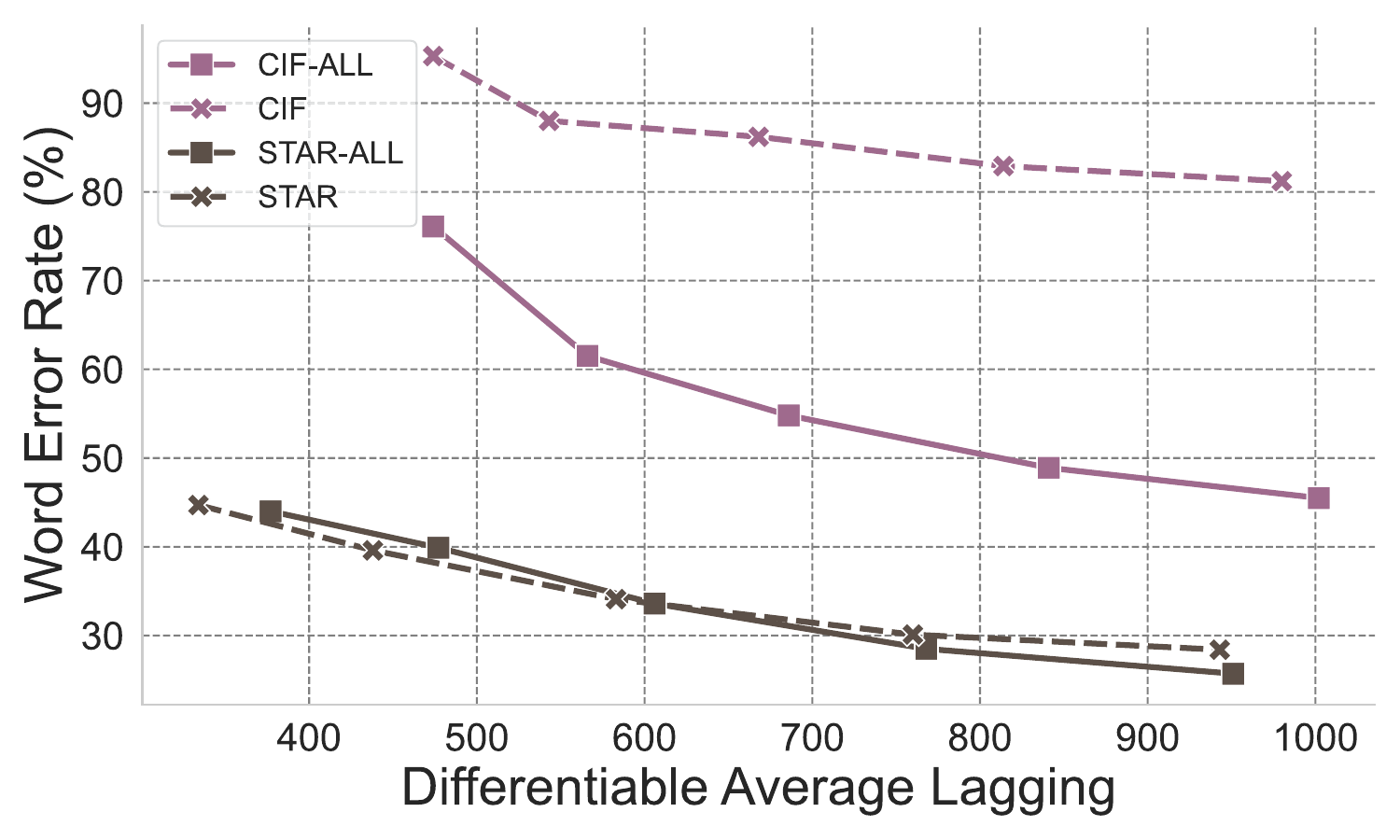}
    \vspace{-1em}
    \caption{Latency-quality trade-off for CIF and \modelname ~using a fixed decision policy instead of their own predicted segmentation. The five markers on the line correspond to five \waitk strategies (from left to right, \waitk $\in \{1,2,3,4,5\}$). }
    \label{figure::fix_compress_simul}
    \vspace{-1.2em}
\end{figure}

\paragraph{Different Segmentations}
In \cref{sec::experiments_stream}, we tested CIF- and \modelname-based models under a shared fixed segmentation policy, where segments were of uniform size ($\lfloor T_x / T_y \rfloor$). This setup evaluates robustness to segmentation changes. Results in \cref{figure::fix_compress_simul} show that while both models experience performance drops, \modelname~ remains robust, achieving $<30$ WER with a DAL of 800, whereas CIF exceeds $80$ WER. This highlights \modelname’s ability to better compress and retain information within anchor representations, making it more robust to policy changes.

Moreover, we let the the two models use all previously computed representations (thus no compression is performed) and name such models CIF-ALL and \modelname-ALL in \cref{figure::fix_compress_simul}. We find that CIF-ALL still greatly lags behind the performance of \modelname ~even when all previous representations are used. This shows that CIF is not a robust method as \textit{it only obtains good performance when aggregating representations using its learned segmentation}. On the contrary, \modelname ~is much more robust; in fact, from \cref{figure::fix_compress_simul}, we find that \modelname ~has a very close performance compared to its non-compressed version \modelname-ALL, providing another evidence of its robust compression quality.

\section{Related work}
\paragraph{End-to-end Streaming Speech-to-text}

For streaming/simultaneous \textit{speech-to-text} tasks, learning speech representation and policies for \readaction and \writeaction is essential. Previous methods like RNN-Transducer \cite{graves2012sequence} and Connectionist Temporal Classification (CTC) \citep{ctc} leverage monotonic alignment for low error rate transcription. Recent work \cite{streaming_asr, tsunoo2020streaming} further extends transformers for streaming ASR using modified attention and beam search.

For speech translation, \citet{ma-etal-2019-stacl} proposed the Wait-K strategy with a fixed decision policy that read chunks of equal-length text for decoding and \citet{ma-etal-2020-simulmt} adapted the wait-k strategy for simultaneous speech translation. Instead of a fixed decision policy, SimulSpeech \cite{ren-etal-2020-simulspeech} trained segmenters with CTC loss. \citet{zeng-etal-2021-realtrans} also use CTC for guidance on word boundary learns to shrink the representation and proposes the Wait-K-Stride-N strategy that writes N tokens for each \readaction action. \citet{dong-etal-2022-learning} and \citet{Chang_2022} use CIF to learn segmentation for the speech sequences and trigger the \writeaction action whenever CIF \fire a new representation. Additionally, \citet{milk} and \citet{Ma2020Monotonic} support a more adaptive strategy where dynamic \readaction and \writeaction are possible. However, even for such an adaptive strategy, a good decision policy still matters \cite{ma-etal-2020-simulmt}.

\paragraph{Efficient Methods for Transformers}
Prior work studied efficient methods to scale Transformers to long sequences~\citep{efficient2022},
including sparse patterns~\citep{longformer2020}, recurrence~\citep{transformerxl19}, kernelized attentions~\citep{performer2021}, etc.
Some of them can be applied in the streaming settings,
such as Streaming LLMs~\citep{xiao2023EfficientStreamingLanguage}, Compressive Transformers~\citep{compressive2020}, etc.
Moreover, \citet{tworkowski2023FocusedTransformerContrastivea,unlimiformer23} proposed to apply $k$NN to the attention to select a subset of past tokens, akin to the segmentation process in this paper.
Similar to the residual connection in our paper, Nugget~\citep{nugget} trains a scorer to select a subset of tokens to represent texts. More recently, \citet{tan-etal-2024-lloco} and \citet{qin2023nugget} also combine context compression with efficient fine-tuning methods like LoRA \cite{hu2021loralowrankadaptationlarge} to expand context length for large language models.

\paragraph{Speech Representation} 
Traditionally, acoustic features are extracted by filter-bank features, mel-frequency cepstral coefficients, or bottleneck features \cite{muda2010voice,mfcc}. More recent work relies on self-supervision to learn speech representations. For example, \citet{zeghidour2021soundstream} and \citet{encodec} learn acoustic representation by reconstructing the original audio. To learn semantic representation, masked language modeling, and contrastive learning objectives are popularized by widely used representations from Hubert \cite{hsu2021hubert}, w2v-BERT \cite{chung2021w2vbert} and Wav2Vec \cite{w2v, w2v2}. All these models use CNNs as a building block to downsample speech signals/representations.

\section{Conclusion and Future Work}
We introduce STAR, a model designed for dynamic compression and transduction of streams. STAR features a segmenter learned via encoder-decoder cross-attention and employs a selection-based compression approach. Our experiments across multiple \textit{speech-to-text} tasks confirm STAR's superior compression performance and latency-quality trade-off relative to established methods such as Convolutional Neural Networks and Continuous Integrate-and-Fire. In the future, we hope to extend this framework to facilitate streaming non-autoregressive generation.

\bibliography{custom}

\begin{thebibliography}{64}
\providecommand{\natexlab}[1]{#1}

\bibitem[{Agarap(2018)}]{Agarap2018DeepLU}
Abien~Fred Agarap. 2018.
\newblock \href {https://api.semanticscholar.org/CorpusID:4090379} {Deep learning using rectified linear units (relu)}.
\newblock \emph{ArXiv}, abs/1803.08375.

\bibitem[{Arivazhagan et~al.(2019)Arivazhagan, Cherry, Macherey, Chiu, Yavuz, Pang, Li, and Raffel}]{milk}
Naveen Arivazhagan, Colin Cherry, Wolfgang Macherey, Chung-Cheng Chiu, Semih Yavuz, Ruoming Pang, Wei Li, and Colin Raffel. 2019.
\newblock \href {https://doi.org/10.18653/v1/P19-1126} {Monotonic infinite lookback attention for simultaneous machine translation}.
\newblock In \emph{Proceedings of the 57th Annual Meeting of the Association for Computational Linguistics}, pages 1313--1323, Florence, Italy. Association for Computational Linguistics.

\bibitem[{Baevski et~al.(2020)Baevski, Zhou, Mohamed, and Auli}]{w2v2}
Alexei Baevski, Henry Zhou, Abdelrahman Mohamed, and Michael Auli. 2020.
\newblock \href {https://arxiv.org/abs/2006.11477} {wav2vec 2.0: A framework for self-supervised learning of speech representations}.
\newblock \emph{Preprint}, arXiv:2006.11477.

\bibitem[{Barrault et~al.(2023)Barrault, Chung, Meglioli, Dale, Dong, Duppenthaler, Duquenne, Ellis, Elsahar, Haaheim, Hoffman, Hwang, Inaguma, Klaiber, Kulikov, Li, Licht, Maillard, Mavlyutov, Rakotoarison, Sadagopan, Ramakrishnan, Tran, Wenzek, Yang, Ye, Evtimov, Fernandez, Gao, Hansanti, Kalbassi, Kallet, Kozhevnikov, Gonzalez, Roman, Touret, Wong, Wood, Yu, Andrews, Balioglu, Chen, Costa-jussà, Elbayad, Gong, Guzmán, Heffernan, Jain, Kao, Lee, Ma, Mourachko, Peloquin, Pino, Popuri, Ropers, Saleem, Schwenk, Sun, Tomasello, Wang, Wang, Wang, and Williamson}]{communication2023seamless}
Loïc Barrault, Yu-An Chung, Mariano~Coria Meglioli, David Dale, Ning Dong, Mark Duppenthaler, Paul-Ambroise Duquenne, Brian Ellis, Hady Elsahar, Justin Haaheim, John Hoffman, Min-Jae Hwang, Hirofumi Inaguma, Christopher Klaiber, Ilia Kulikov, Pengwei Li, Daniel Licht, Jean Maillard, Ruslan Mavlyutov, Alice Rakotoarison, Kaushik~Ram Sadagopan, Abinesh Ramakrishnan, Tuan Tran, Guillaume Wenzek, Yilin Yang, Ethan Ye, Ivan Evtimov, Pierre Fernandez, Cynthia Gao, Prangthip Hansanti, Elahe Kalbassi, Amanda Kallet, Artyom Kozhevnikov, Gabriel~Mejia Gonzalez, Robin~San Roman, Christophe Touret, Corinne Wong, Carleigh Wood, Bokai Yu, Pierre Andrews, Can Balioglu, Peng-Jen Chen, Marta~R. Costa-jussà, Maha Elbayad, Hongyu Gong, Francisco Guzmán, Kevin Heffernan, Somya Jain, Justine Kao, Ann Lee, Xutai Ma, Alex Mourachko, Benjamin Peloquin, Juan Pino, Sravya Popuri, Christophe Ropers, Safiyyah Saleem, Holger Schwenk, Anna Sun, Paden Tomasello, Changhan Wang, Jeff Wang, Skyler Wang, and Mary Williamson. 2023.
\newblock \href {https://arxiv.org/abs/2312.05187} {Seamless: Multilingual expressive and streaming speech translation}.
\newblock \emph{Preprint}, arXiv:2312.05187.

\bibitem[{Beltagy et~al.(2020)Beltagy, Peters, and Cohan}]{longformer2020}
Iz~Beltagy, Matthew~E. Peters, and Arman Cohan. 2020.
\newblock \href {http://arxiv.org/abs/2004.05150} {Longformer: {{The Long-Document Transformer}}}.

\bibitem[{Bertsch et~al.(2023)Bertsch, Alon, Neubig, and Gormley}]{unlimiformer23}
Amanda Bertsch, Uri Alon, Graham Neubig, and Matthew~R. Gormley. 2023.
\newblock \href {http://arxiv.org/abs/2305.01625} {Unlimiformer: {{Long-Range Transformers}} with {{Unlimited Length Input}}}.

\bibitem[{Chang and Lee(2022)}]{Chang_2022}
Chih-Chiang Chang and Hung-yi Lee. 2022.
\newblock \href {https://doi.org/10.21437/interspeech.2022-10627} {Exploring continuous integrate-and-fire for adaptive simultaneous speech translation}.
\newblock In \emph{Interspeech 2022}. ISCA.

\bibitem[{Chen et~al.(2021)Chen, Ma, Zheng, and Huang}]{Chen2021DirectSS}
Junkun Chen, Mingbo Ma, Renjie Zheng, and Liang Huang. 2021.
\newblock \href {https://api.semanticscholar.org/CorpusID:235422036} {Direct simultaneous speech-to-text translation assisted by synchronized streaming asr}.
\newblock In \emph{Findings}.

\bibitem[{Choromanski et~al.(2021)Choromanski, Likhosherstov, Dohan, Song, Gane, Sarlos, Hawkins, Davis, Mohiuddin, Kaiser, Belanger, Colwell, and Weller}]{performer2021}
Krzysztof Choromanski, Valerii Likhosherstov, David Dohan, Xingyou Song, Andreea Gane, Tamas Sarlos, Peter Hawkins, Jared Davis, Afroz Mohiuddin, Lukasz Kaiser, David Belanger, Lucy Colwell, and Adrian Weller. 2021.
\newblock \href {http://arxiv.org/abs/2009.14794} {Rethinking {{Attention}} with {{Performers}}}.
\newblock In \emph{International Conference on Learning Representations (ICLR)}.

\bibitem[{Chung et~al.(2014)Chung, Gulcehre, Cho, and Bengio}]{chung2014empirical}
Junyoung Chung, Caglar Gulcehre, KyungHyun Cho, and Yoshua Bengio. 2014.
\newblock \href {https://arxiv.org/abs/1412.3555} {Empirical evaluation of gated recurrent neural networks on sequence modeling}.
\newblock \emph{Preprint}, arXiv:1412.3555.

\bibitem[{Chung et~al.(2021)Chung, Zhang, Han, Chiu, Qin, Pang, and Wu}]{chung2021w2vbert}
Yu-An Chung, Yu~Zhang, Wei Han, Chung-Cheng Chiu, James Qin, Ruoming Pang, and Yonghui Wu. 2021.
\newblock \href {https://arxiv.org/abs/2108.06209} {W2v-bert: Combining contrastive learning and masked language modeling for self-supervised speech pre-training}.
\newblock \emph{Preprint}, arXiv:2108.06209.

\bibitem[{Dai et~al.(2019)Dai, Yang, Yang, Carbonell, Le, and Salakhutdinov}]{transformerxl19}
Zihang Dai, Zhilin Yang, Yiming Yang, Jaime Carbonell, Quoc~V. Le, and Ruslan Salakhutdinov. 2019.
\newblock \href {http://arxiv.org/abs/1901.02860} {Transformer-{{XL}}: {{Attentive Language Models Beyond}} a {{Fixed-Length Context}}}.
\newblock In \emph{Annual Meeting of the Association for Computational Linguistics (ACL)}.

\bibitem[{Davis and Mermelstein(1980)}]{mfcc}
S.~Davis and P.~Mermelstein. 1980.
\newblock \href {https://doi.org/10.1109/TASSP.1980.1163420} {Comparison of parametric representations for monosyllabic word recognition in continuously spoken sentences}.
\newblock \emph{IEEE Transactions on Acoustics, Speech, and Signal Processing}, 28(4):357--366.

\bibitem[{Devlin et~al.(2019)Devlin, Chang, Lee, and Toutanova}]{Devlin2019BERTPO}
Jacob Devlin, Ming-Wei Chang, Kenton Lee, and Kristina Toutanova. 2019.
\newblock \href {https://api.semanticscholar.org/CorpusID:52967399} {Bert: Pre-training of deep bidirectional transformers for language understanding}.
\newblock In \emph{North American Chapter of the Association for Computational Linguistics}.

\bibitem[{Di~Gangi et~al.(2019)Di~Gangi, Cattoni, Bentivogli, Negri, and Turchi}]{di-gangi-etal-2019-must}
Mattia~A. Di~Gangi, Roldano Cattoni, Luisa Bentivogli, Matteo Negri, and Marco Turchi. 2019.
\newblock \href {https://doi.org/10.18653/v1/N19-1202} {{M}u{ST}-{C}: a {M}ultilingual {S}peech {T}ranslation {C}orpus}.
\newblock In \emph{Proceedings of the 2019 Conference of the North {A}merican Chapter of the Association for Computational Linguistics: Human Language Technologies, Volume 1 (Long and Short Papers)}, pages 2012--2017, Minneapolis, Minnesota. Association for Computational Linguistics.

\bibitem[{Dong and Xu(2020)}]{dong2020cif}
Linhao Dong and Bo~Xu. 2020.
\newblock \href {https://arxiv.org/abs/1905.11235} {Cif: Continuous integrate-and-fire for end-to-end speech recognition}.
\newblock \emph{Preprint}, arXiv:1905.11235.

\bibitem[{Dong et~al.(2022)Dong, Zhu, Wang, and Li}]{dong-etal-2022-learning}
Qian Dong, Yaoming Zhu, Mingxuan Wang, and Lei Li. 2022.
\newblock \href {https://doi.org/10.18653/v1/2022.acl-long.50} {Learning when to translate for streaming speech}.
\newblock In \emph{Proceedings of the 60th Annual Meeting of the Association for Computational Linguistics (Volume 1: Long Papers)}, pages 680--694, Dublin, Ireland. Association for Computational Linguistics.

\bibitem[{Défossez et~al.(2022)Défossez, Copet, Synnaeve, and Adi}]{encodec}
Alexandre Défossez, Jade Copet, Gabriel Synnaeve, and Yossi Adi. 2022.
\newblock \href {https://arxiv.org/abs/2210.13438} {High fidelity neural audio compression}.
\newblock \emph{Preprint}, arXiv:2210.13438.

\bibitem[{Fonseca et~al.(2017)Fonseca, Pons, Favory, Font, Bogdanov, Ferraro, Oramas, Porter, and Serra}]{freesound}
Eduardo Fonseca, Jordi Pons, Xavier Favory, Frederic Font, Dmitry Bogdanov, Andres Ferraro, Sergio Oramas, Alastair Porter, and Xavier Serra. 2017.
\newblock Freesound datasets: A platform for the creation of open audio datasets.

\bibitem[{Graves(2012)}]{graves2012sequence}
Alex Graves. 2012.
\newblock \href {https://arxiv.org/abs/1211.3711} {Sequence transduction with recurrent neural networks}.
\newblock \emph{Preprint}, arXiv:1211.3711.

\bibitem[{Graves et~al.(2006)Graves, Fern\'{a}ndez, Gomez, and Schmidhuber}]{ctc}
Alex Graves, Santiago Fern\'{a}ndez, Faustino Gomez, and J\"{u}rgen Schmidhuber. 2006.
\newblock \href {https://doi.org/10.1145/1143844.1143891} {Connectionist temporal classification: labelling unsegmented sequence data with recurrent neural networks}.
\newblock In \emph{Proceedings of the 23rd International Conference on Machine Learning}, ICML '06, page 369–376, New York, NY, USA. Association for Computing Machinery.

\bibitem[{Gulati et~al.(2020)Gulati, Qin, Chiu, Parmar, Zhang, Yu, Han, Wang, Zhang, Wu, and Pang}]{conformer}
Anmol Gulati, James Qin, Chung{-}Cheng Chiu, Niki Parmar, Yu~Zhang, Jiahui Yu, Wei Han, Shibo Wang, Zhengdong Zhang, Yonghui Wu, and Ruoming Pang. 2020.
\newblock \href {https://arxiv.org/abs/2005.08100} {Conformer: Convolution-augmented transformer for speech recognition}.
\newblock \emph{CoRR}, abs/2005.08100.

\bibitem[{Hochreiter and Schmidhuber(1997)}]{HochSchm97}
Sepp Hochreiter and Jürgen Schmidhuber. 1997.
\newblock Long short-term memory.
\newblock \emph{Neural Computation}, 9(8):1735--1780.

\bibitem[{Hsu et~al.(2021)Hsu, Bolte, Tsai, Lakhotia, Salakhutdinov, and Mohamed}]{hsu2021hubert}
Wei-Ning Hsu, Benjamin Bolte, Yao-Hung~Hubert Tsai, Kushal Lakhotia, Ruslan Salakhutdinov, and Abdelrahman Mohamed. 2021.
\newblock \href {https://arxiv.org/abs/2106.07447} {Hubert: Self-supervised speech representation learning by masked prediction of hidden units}.
\newblock \emph{Preprint}, arXiv:2106.07447.

\bibitem[{Hu et~al.(2021)Hu, Shen, Wallis, Allen-Zhu, Li, Wang, Wang, and Chen}]{hu2021loralowrankadaptationlarge}
Edward~J. Hu, Yelong Shen, Phillip Wallis, Zeyuan Allen-Zhu, Yuanzhi Li, Shean Wang, Lu~Wang, and Weizhu Chen. 2021.
\newblock \href {https://arxiv.org/abs/2106.09685} {Lora: Low-rank adaptation of large language models}.
\newblock \emph{Preprint}, arXiv:2106.09685.

\bibitem[{Inaguma et~al.(2020)Inaguma, Gaur, Lu, Li, and Gong}]{Inaguma2020MinimumLT}
Hirofumi Inaguma, Yashesh Gaur, Liang Lu, Jinyu Li, and Yifan Gong. 2020.
\newblock \href {https://api.semanticscholar.org/CorpusID:215737044} {Minimum latency training strategies for streaming sequence-to-sequence asr}.
\newblock \emph{ICASSP 2020 - 2020 IEEE International Conference on Acoustics, Speech and Signal Processing (ICASSP)}, pages 6064--6068.

\bibitem[{Kameoka et~al.(2021)Kameoka, Tanaka, and Kaneko}]{Kameoka2021FastS2SVCSN}
H.~Kameoka, Kou Tanaka, and Takuhiro Kaneko. 2021.
\newblock \href {https://api.semanticscholar.org/CorpusID:233231613} {Fasts2s-vc: Streaming non-autoregressive sequence-to-sequence voice conversion}.
\newblock \emph{ArXiv}, abs/2104.06900.

\bibitem[{Khattab and Zaharia(2020)}]{Khattab2020ColBERTEA}
O.~Khattab and Matei~A. Zaharia. 2020.
\newblock \href {https://api.semanticscholar.org/CorpusID:216553223} {Colbert: Efficient and effective passage search via contextualized late interaction over bert}.
\newblock \emph{Proceedings of the 43rd International ACM SIGIR Conference on Research and Development in Information Retrieval}.

\bibitem[{Krizhevsky et~al.(2012)Krizhevsky, Sutskever, and Hinton}]{conv}
Alex Krizhevsky, Ilya Sutskever, and Geoffrey~E Hinton. 2012.
\newblock \href {https://proceedings.neurips.cc/paper_files/paper/2012/file/c399862d3b9d6b76c8436e924a68c45b-Paper.pdf} {Imagenet classification with deep convolutional neural networks}.
\newblock In \emph{Advances in Neural Information Processing Systems}, volume~25. Curran Associates, Inc.

\bibitem[{Lecun and Bengio(1995)}]{cnn}
Yann Lecun and Yoshua Bengio. 1995.
\newblock \emph{Convolutional Networks for Images, Speech and Time Series}, pages 255--258.
\newblock The MIT Press.

\bibitem[{Li(2021)}]{Li2021RecentAI}
Jinyu Li. 2021.
\newblock \href {https://api.semanticscholar.org/CorpusID:240419899} {Recent advances in end-to-end automatic speech recognition}.
\newblock \emph{ArXiv}, abs/2111.01690.

\bibitem[{Li et~al.(2020)Li, Wang, Tang, Tran, Tang, Pino, Baevski, Conneau, and Auli}]{Li2020MultilingualST}
Xian Li, Changhan Wang, Yun Tang, C.~Tran, Yuqing Tang, Juan~Miguel Pino, Alexei Baevski, Alexis Conneau, and Michael Auli. 2020.
\newblock \href {https://api.semanticscholar.org/CorpusID:227840539} {Multilingual speech translation from efficient finetuning of pretrained models}.
\newblock In \emph{Annual Meeting of the Association for Computational Linguistics}.

\bibitem[{Lipton(2015)}]{rnn_survey}
Zachary~Chase Lipton. 2015.
\newblock \href {https://arxiv.org/abs/1506.00019} {A critical review of recurrent neural networks for sequence learning}.
\newblock \emph{CoRR}, abs/1506.00019.

\bibitem[{Liu et~al.(2021)Liu, Du, Li, Li, and Chen}]{liu-etal-2021-cross}
Dan Liu, Mengge Du, Xiaoxi Li, Ya~Li, and Enhong Chen. 2021.
\newblock \href {https://doi.org/10.18653/v1/2021.emnlp-main.4} {Cross attention augmented transducer networks for simultaneous translation}.
\newblock In \emph{Proceedings of the 2021 Conference on Empirical Methods in Natural Language Processing}, pages 39--55, Online and Punta Cana, Dominican Republic. Association for Computational Linguistics.

\bibitem[{Liu et~al.(2019)Liu, Xiong, He, Zhang, Wu, Wang, and Zong}]{Liu2019EndtoEndST}
Yuchen Liu, Hao Xiong, Zhongjun He, Jiajun Zhang, Hua Wu, Haifeng Wang, and Chengqing Zong. 2019.
\newblock \href {https://api.semanticscholar.org/CorpusID:119309065} {End-to-end speech translation with knowledge distillation}.
\newblock In \emph{Interspeech}.

\bibitem[{Luong et~al.(2015)Luong, Pham, and Manning}]{luong-etal-2015-effective}
Thang Luong, Hieu Pham, and Christopher~D. Manning. 2015.
\newblock \href {https://doi.org/10.18653/v1/D15-1166} {Effective approaches to attention-based neural machine translation}.
\newblock In \emph{Proceedings of the 2015 Conference on Empirical Methods in Natural Language Processing}, pages 1412--1421, Lisbon, Portugal. Association for Computational Linguistics.

\bibitem[{Ma et~al.(2019)Ma, Huang, Xiong, Zheng, Liu, Zheng, Zhang, He, Liu, Li, Wu, and Wang}]{ma-etal-2019-stacl}
Mingbo Ma, Liang Huang, Hao Xiong, Renjie Zheng, Kaibo Liu, Baigong Zheng, Chuanqiang Zhang, Zhongjun He, Hairong Liu, Xing Li, Hua Wu, and Haifeng Wang. 2019.
\newblock \href {https://doi.org/10.18653/v1/P19-1289} {{STACL}: Simultaneous translation with implicit anticipation and controllable latency using prefix-to-prefix framework}.
\newblock In \emph{Proceedings of the 57th Annual Meeting of the Association for Computational Linguistics}, pages 3025--3036, Florence, Italy. Association for Computational Linguistics.

\bibitem[{Ma et~al.(2020{\natexlab{a}})Ma, Dousti, Wang, Gu, and Pino}]{ma-etal-2020-simuleval}
Xutai Ma, Mohammad~Javad Dousti, Changhan Wang, Jiatao Gu, and Juan Pino. 2020{\natexlab{a}}.
\newblock \href {https://doi.org/10.18653/v1/2020.emnlp-demos.19} {{SIMULEVAL}: An evaluation toolkit for simultaneous translation}.
\newblock In \emph{Proceedings of the 2020 Conference on Empirical Methods in Natural Language Processing: System Demonstrations}, pages 144--150, Online. Association for Computational Linguistics.

\bibitem[{Ma et~al.(2020{\natexlab{b}})Ma, Pino, and Koehn}]{ma-etal-2020-simulmt}
Xutai Ma, Juan Pino, and Philipp Koehn. 2020{\natexlab{b}}.
\newblock \href {https://aclanthology.org/2020.aacl-main.58} {{S}imul{MT} to {S}imul{ST}: Adapting simultaneous text translation to end-to-end simultaneous speech translation}.
\newblock In \emph{Proceedings of the 1st Conference of the Asia-Pacific Chapter of the Association for Computational Linguistics and the 10th International Joint Conference on Natural Language Processing}, pages 582--587, Suzhou, China. Association for Computational Linguistics.

\bibitem[{Ma et~al.(2020{\natexlab{c}})Ma, Pino, Cross, Puzon, and Gu}]{Ma2020Monotonic}
Xutai Ma, Juan~Miguel Pino, James Cross, Liezl Puzon, and Jiatao Gu. 2020{\natexlab{c}}.
\newblock \href {https://openreview.net/forum?id=Hyg96gBKPS} {Monotonic multihead attention}.
\newblock In \emph{International Conference on Learning Representations}.

\bibitem[{Moritz et~al.(2020)Moritz, Hori, and Le}]{streaming_asr}
Niko Moritz, Takaaki Hori, and Jonathan Le. 2020.
\newblock \href {https://doi.org/10.1109/ICASSP40776.2020.9054476} {Streaming automatic speech recognition with the transformer model}.
\newblock In \emph{ICASSP 2020 - 2020 IEEE International Conference on Acoustics, Speech and Signal Processing (ICASSP)}, pages 6074--6078.

\bibitem[{Morris et~al.(2004)Morris, Maier, and Green}]{wer}
Andrew~C. Morris, Viktoria Maier, and Phil~D. Green. 2004.
\newblock \href {https://api.semanticscholar.org/CorpusID:18880375} {From wer and ril to mer and wil: improved evaluation measures for connected speech recognition}.
\newblock In \emph{Interspeech}.

\bibitem[{Muda et~al.(2010)Muda, Begam, and Elamvazuthi}]{muda2010voice}
Lindasalwa Muda, Mumtaj Begam, and I.~Elamvazuthi. 2010.
\newblock \href {https://arxiv.org/abs/1003.4083} {Voice recognition algorithms using mel frequency cepstral coefficient (mfcc) and dynamic time warping (dtw) techniques}.
\newblock \emph{Preprint}, arXiv:1003.4083.

\bibitem[{Panayotov et~al.(2015)Panayotov, Chen, Povey, and Khudanpur}]{librisppech_dataset}
Vassil Panayotov, Guoguo Chen, Daniel Povey, and Sanjeev Khudanpur. 2015.
\newblock \href {https://doi.org/10.1109/ICASSP.2015.7178964} {Librispeech: An asr corpus based on public domain audio books}.
\newblock In \emph{2015 IEEE International Conference on Acoustics, Speech and Signal Processing (ICASSP)}, pages 5206--5210.

\bibitem[{Papineni et~al.(2002)Papineni, Roukos, Ward, and Zhu}]{papineni-etal-2002-bleu}
Kishore Papineni, Salim Roukos, Todd Ward, and Wei-Jing Zhu. 2002.
\newblock \href {https://doi.org/10.3115/1073083.1073135} {{B}leu: a method for automatic evaluation of machine translation}.
\newblock In \emph{Proceedings of the 40th Annual Meeting of the Association for Computational Linguistics}, pages 311--318, Philadelphia, Pennsylvania, USA. Association for Computational Linguistics.

\bibitem[{Prabhavalkar et~al.(2023)Prabhavalkar, Hori, Sainath, Schlüter, and Watanabe}]{prabhavalkar2023endtoend}
Rohit Prabhavalkar, Takaaki Hori, Tara~N. Sainath, Ralf Schlüter, and Shinji Watanabe. 2023.
\newblock \href {https://arxiv.org/abs/2303.03329} {End-to-end speech recognition: A survey}.
\newblock \emph{Preprint}, arXiv:2303.03329.

\bibitem[{Qin et~al.(2023)Qin, Rosset, Chau, Rao, and Durme}]{qin2023nugget}
Guanghui Qin, Corby Rosset, Ethan~C. Chau, Nikhil Rao, and Benjamin~Van Durme. 2023.
\newblock \href {https://arxiv.org/abs/2310.02409} {Nugget 2d: Dynamic contextual compression for scaling decoder-only language models}.
\newblock \emph{Preprint}, arXiv:2310.02409.

\bibitem[{Qin and Van~Durme(2023)}]{nugget}
Guanghui Qin and Benjamin Van~Durme. 2023.
\newblock \href {https://proceedings.mlr.press/v202/qin23a.html} {Nugget: Neural agglomerative embeddings of text}.
\newblock In \emph{Proceedings of the 40th International Conference on Machine Learning}, volume 202 of \emph{Proceedings of Machine Learning Research}, pages 28337--28350. PMLR.

\bibitem[{Rae et~al.(2020)Rae, Potapenko, Jayakumar, and Lillicrap}]{compressive2020}
Jack~W. Rae, Anna Potapenko, Siddhant~M. Jayakumar, and Timothy~P. Lillicrap. 2020.
\newblock \href {http://arxiv.org/abs/1911.05507} {Compressive {{Transformers}} for {{Long-Range Sequence Modelling}}}.
\newblock In \emph{International Conference on Learning Representations (ICLR)}.

\bibitem[{Ren et~al.(2020)Ren, Liu, Tan, Zhang, Qin, Zhao, and Liu}]{ren-etal-2020-simulspeech}
Yi~Ren, Jinglin Liu, Xu~Tan, Chen Zhang, Tao Qin, Zhou Zhao, and Tie-Yan Liu. 2020.
\newblock \href {https://doi.org/10.18653/v1/2020.acl-main.350} {{S}imul{S}peech: End-to-end simultaneous speech to text translation}.
\newblock In \emph{Proceedings of the 58th Annual Meeting of the Association for Computational Linguistics}, pages 3787--3796, Online. Association for Computational Linguistics.

\bibitem[{Schneider et~al.(2019)Schneider, Baevski, Collobert, and Auli}]{w2v}
Steffen Schneider, Alexei Baevski, Ronan Collobert, and Michael Auli. 2019.
\newblock \href {https://arxiv.org/abs/1904.05862} {wav2vec: Unsupervised pre-training for speech recognition}.
\newblock \emph{Preprint}, arXiv:1904.05862.

\bibitem[{Sennrich et~al.(2016)Sennrich, Haddow, and Birch}]{bpe}
Rico Sennrich, Barry Haddow, and Alexandra Birch. 2016.
\newblock \href {https://doi.org/10.18653/v1/P16-1162} {Neural machine translation of rare words with subword units}.
\newblock In \emph{Proceedings of the 54th Annual Meeting of the Association for Computational Linguistics (Volume 1: Long Papers)}, pages 1715--1725, Berlin, Germany. Association for Computational Linguistics.

\bibitem[{Tan et~al.(2024)Tan, Li, Patil, Wu, Zhang, Keutzer, Gonzalez, and Popa}]{tan-etal-2024-lloco}
Sijun Tan, Xiuyu Li, Shishir~G Patil, Ziyang Wu, Tianjun Zhang, Kurt Keutzer, Joseph~E. Gonzalez, and Raluca~Ada Popa. 2024.
\newblock \href {https://doi.org/10.18653/v1/2024.emnlp-main.975} {{LL}o{CO}: Learning long contexts offline}.
\newblock In \emph{Proceedings of the 2024 Conference on Empirical Methods in Natural Language Processing}, pages 17605--17621, Miami, Florida, USA. Association for Computational Linguistics.

\bibitem[{Tay et~al.(2022)Tay, Dehghani, Bahri, and Metzler}]{efficient2022}
Yi~Tay, Mostafa Dehghani, Dara Bahri, and Donald Metzler. 2022.
\newblock \href {http://arxiv.org/abs/2009.06732} {Efficient {{Transformers}}: {{A Survey}}}.
\newblock \emph{ACM Computing Surveys}, 55(6):1--28.

\bibitem[{Tsunoo et~al.(2020)Tsunoo, Kashiwagi, and Watanabe}]{tsunoo2020streaming}
Emiru Tsunoo, Yosuke Kashiwagi, and Shinji Watanabe. 2020.
\newblock \href {https://arxiv.org/abs/2006.14941} {Streaming transformer asr with blockwise synchronous beam search}.
\newblock \emph{Preprint}, arXiv:2006.14941.

\bibitem[{Tworkowski et~al.(2023)Tworkowski, Staniszewski, Pacek, Wu, Michalewski, and Miłoś}]{tworkowski2023FocusedTransformerContrastivea}
Szymon Tworkowski, Konrad Staniszewski, Mikołaj Pacek, Yuhuai Wu, Henryk Michalewski, and Piotr Miłoś. 2023.
\newblock \href {http://arxiv.org/abs/2307.03170} {Focused {{Transformer}}: {{Contrastive Training}} for {{Context Scaling}}}.

\bibitem[{Vaswani et~al.(2017)Vaswani, Shazeer, Parmar, Uszkoreit, Jones, Gomez, Kaiser, and Polosukhin}]{transformer}
Ashish Vaswani, Noam Shazeer, Niki Parmar, Jakob Uszkoreit, Llion Jones, Aidan~N Gomez, \L~ukasz Kaiser, and Illia Polosukhin. 2017.
\newblock \href {https://proceedings.neurips.cc/paper_files/paper/2017/file/3f5ee243547dee91fbd053c1c4a845aa-Paper.pdf} {Attention is all you need}.
\newblock In \emph{Advances in Neural Information Processing Systems}, volume~30. Curran Associates, Inc.

\bibitem[{Wang et~al.(2022)Wang, Sun, Xue, Wu, Zhou, Gaur, Liu, and Li}]{Wang2022LAMASSUAS}
Peidong Wang, Eric Sun, Jian Xue, Yu~Wu, Long Zhou, Yashesh Gaur, Shujie Liu, and Jinyu Li. 2022.
\newblock \href {https://api.semanticscholar.org/CorpusID:258968116} {Lamassu: A streaming language-agnostic multilingual speech recognition and translation model using neural transducers}.
\newblock \emph{INTERSPEECH 2023}.

\bibitem[{Xiao et~al.(2023)Xiao, Tian, Chen, Han, and Lewis}]{xiao2023EfficientStreamingLanguage}
Guangxuan Xiao, Yuandong Tian, Beidi Chen, Song Han, and Mike Lewis. 2023.
\newblock \href {http://arxiv.org/abs/2309.17453} {Efficient {{Streaming Language Models}} with {{Attention Sinks}}}.

\bibitem[{Xu et~al.(2015)Xu, Ba, Kiros, Cho, Courville, Salakhudinov, Zemel, and Bengio}]{sft_hard_attn}
Kelvin Xu, Jimmy Ba, Ryan Kiros, Kyunghyun Cho, Aaron Courville, Ruslan Salakhudinov, Rich Zemel, and Yoshua Bengio. 2015.
\newblock \href {https://proceedings.mlr.press/v37/xuc15.html} {Show, attend and tell: Neural image caption generation with visual attention}.
\newblock In \emph{Proceedings of the 32nd International Conference on Machine Learning}, volume~37 of \emph{Proceedings of Machine Learning Research}, pages 2048--2057, Lille, France. PMLR.

\bibitem[{Xue et~al.(2022)Xue, Wang, Li, Post, and Gaur}]{Xue2022LargeScaleSE}
Jian Xue, Peidong Wang, Jinyu Li, Matt Post, and Yashesh Gaur. 2022.
\newblock \href {https://api.semanticscholar.org/CorpusID:248118691} {Large-scale streaming end-to-end speech translation with neural transducers}.
\newblock In \emph{Interspeech}.

\bibitem[{Zeghidour et~al.(2021)Zeghidour, Luebs, Omran, Skoglund, and Tagliasacchi}]{zeghidour2021soundstream}
Neil Zeghidour, Alejandro Luebs, Ahmed Omran, Jan Skoglund, and Marco Tagliasacchi. 2021.
\newblock \href {https://arxiv.org/abs/2107.03312} {Soundstream: An end-to-end neural audio codec}.
\newblock \emph{Preprint}, arXiv:2107.03312.

\bibitem[{Zen et~al.(2019)Zen, Dang, Clark, Zhang, Weiss, Jia, Chen, and Wu}]{zen2019libritts}
Heiga Zen, Viet Dang, Rob Clark, Yu~Zhang, Ron~J. Weiss, Ye~Jia, Zhifeng Chen, and Yonghui Wu. 2019.
\newblock \href {https://arxiv.org/abs/1904.02882} {Libritts: A corpus derived from librispeech for text-to-speech}.
\newblock \emph{Preprint}, arXiv:1904.02882.

\bibitem[{Zeng et~al.(2021)Zeng, Li, and Liu}]{zeng-etal-2021-realtrans}
Xingshan Zeng, Liangyou Li, and Qun Liu. 2021.
\newblock \href {https://doi.org/10.18653/v1/2021.findings-acl.218} {{R}eal{T}ran{S}: End-to-end simultaneous speech translation with convolutional weighted-shrinking transformer}.
\newblock In \emph{Findings of the Association for Computational Linguistics: ACL-IJCNLP 2021}, pages 2461--2474, Online. Association for Computational Linguistics.

\end{thebibliography}

\onecolumn
\section*{\LARGE{Supplementary Material}}
\appendix

\begin{table}[ht]
    \centering
    \footnotesize
    \begin{tabular}{cl}
    \textbf{Appendix Sections}    & \textbf{Contents}  \\ \toprule
    \autoref{app::segmentation_analysis} &  \begin{tabular}[c]{@{}l@{}} Qualitative Examples of Speech Segmentation \end{tabular} \\ \midrule
    \autoref{app::cif} &  \begin{tabular}[c]{@{}l@{}} More Details on Continuous Integrate and Fire \end{tabular} \\ \midrule
    \autoref{app::noise_injection} &  \begin{tabular}[c]{@{}l@{}} STAR's Robustness to Noise Injection \end{tabular} \\ \midrule
     \autoref{app::sts} &  \begin{tabular}[c]{@{}l@{}} Similarity Test for Compressed Speech Representation  \end{tabular} \\ \midrule
   \autoref{app::memory}     &  \begin{tabular}[c]{@{}l@{}} Benchmark Memory Usage with/without Compression\end{tabular} \\ \midrule
    \autoref{app::hyper}     &  \begin{tabular}[c]{@{}l@{}} Model Configurations and Hyper-parameters\end{tabular} \\ \midrule
     \autoref{app::dal}     &  \begin{tabular}[c]{@{}l@{}} Measuring Latency: Differentiable Average Lagging\end{tabular} \\ \midrule
\end{tabular}    
\end{table}

\section{Qualitative Examples of Speech Segmentation from Compressors}\label{app::segmentation_analysis}

\begin{figure*}[h!]
    \centering
    \includegraphics[width=0.85\linewidth]{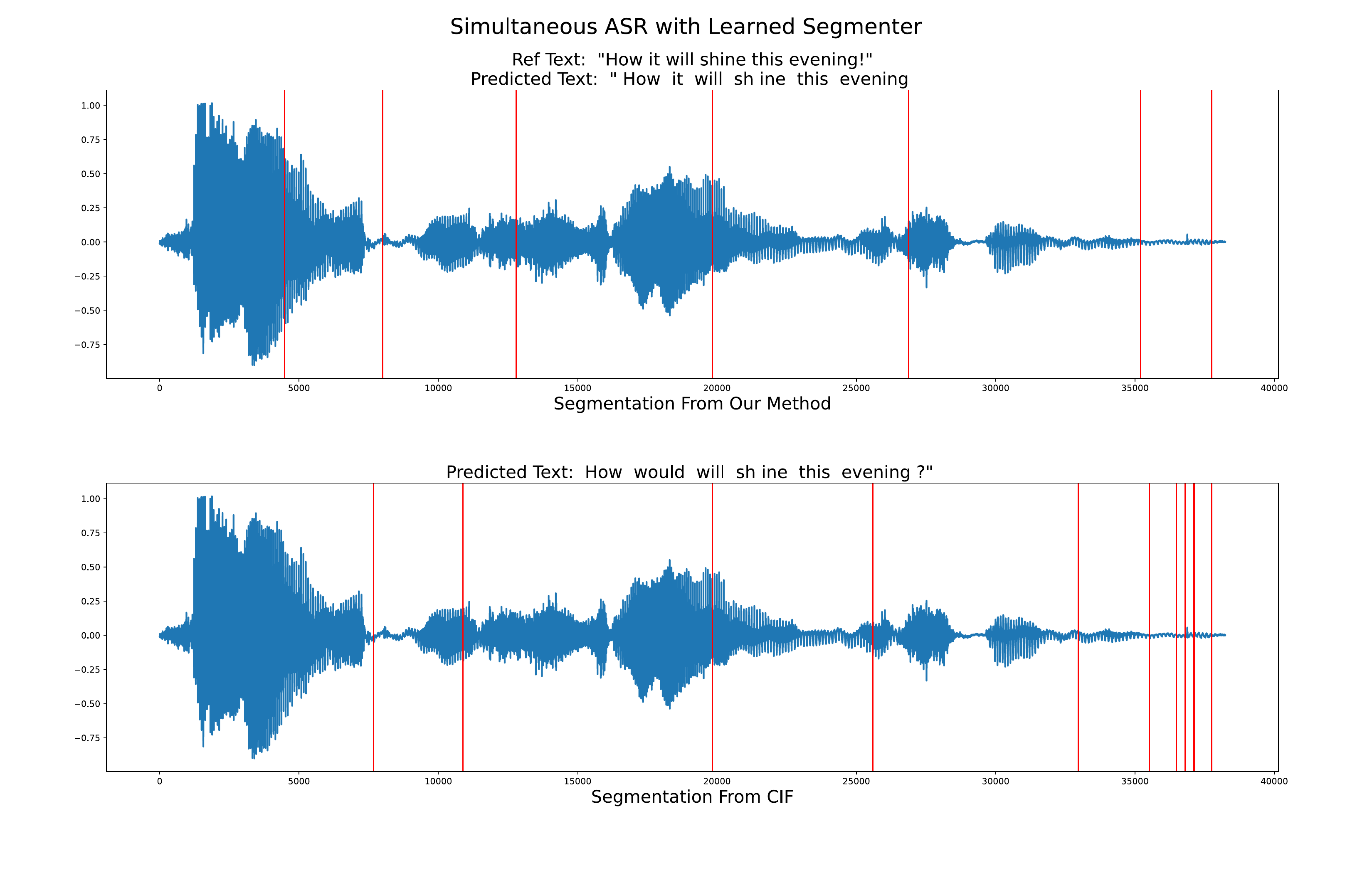}
     \caption{Qualitative Examples of CIF and \modelname ~based Segmentation for Simul ASR}.
    \label{figure::qual_1}
\end{figure*}

\begin{figure*}[h!]
    \centering
    \includegraphics[width=0.85\linewidth]{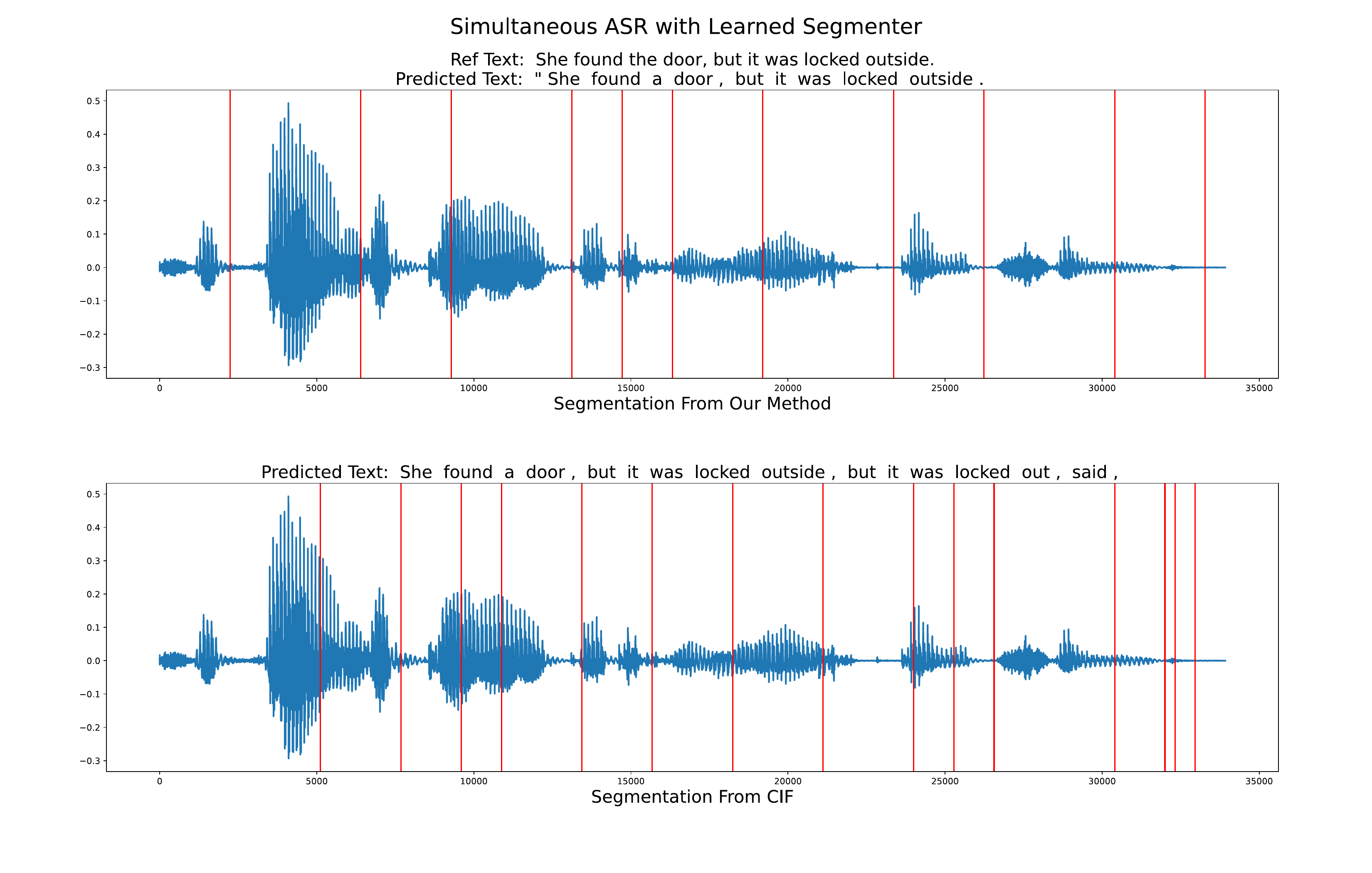}
    \includegraphics[width=0.85\linewidth]{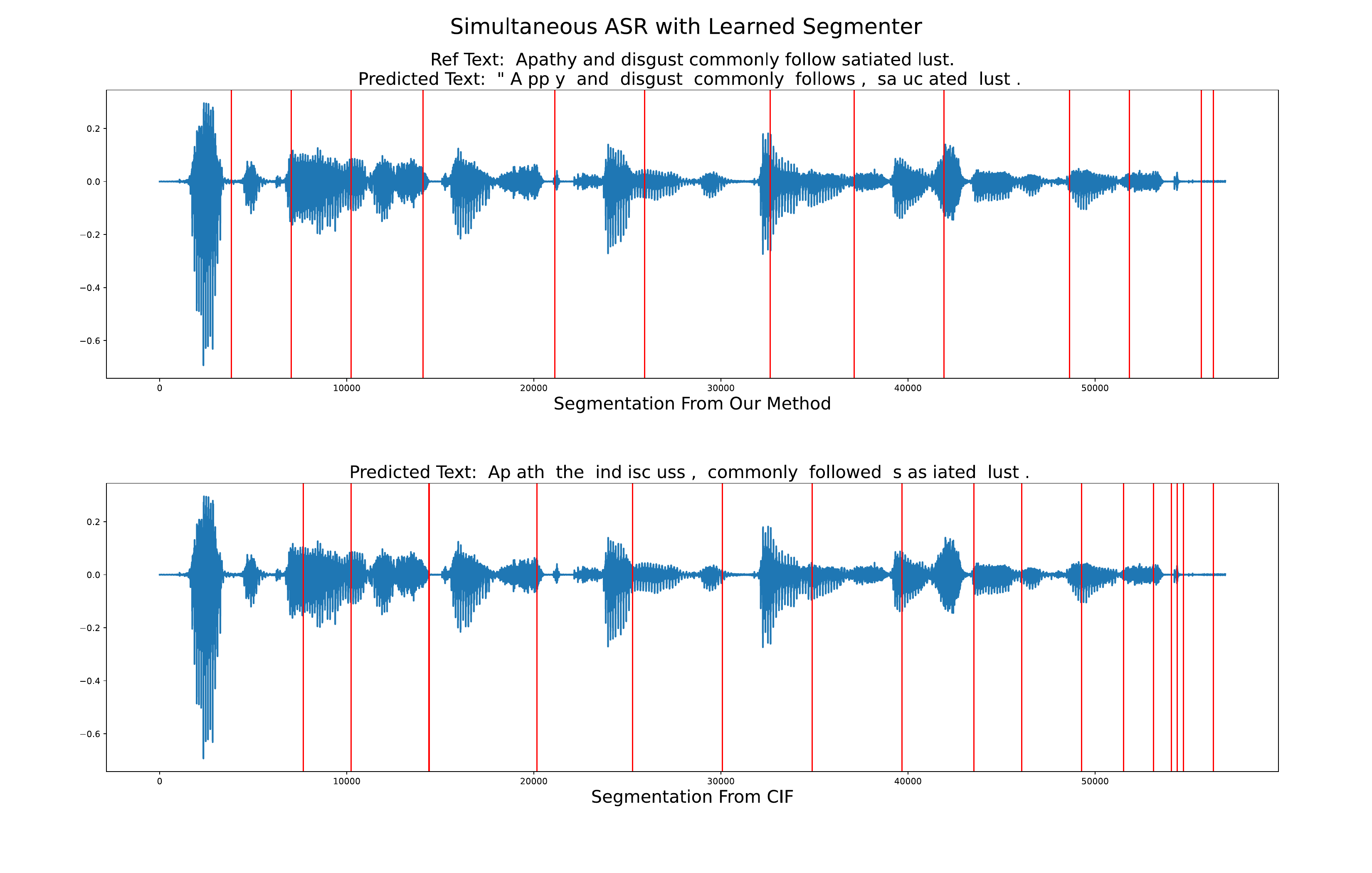}
     \caption{Qualitative Examples of CIF and \modelname ~based Segmentation for Simul ASR}.
    \label{figure::qual_2}
\end{figure*}
\clearpage

\section{Continuous Integrate and Fire}\label{app::cif}

\begin{figure}[h]
    \centering
    \includegraphics{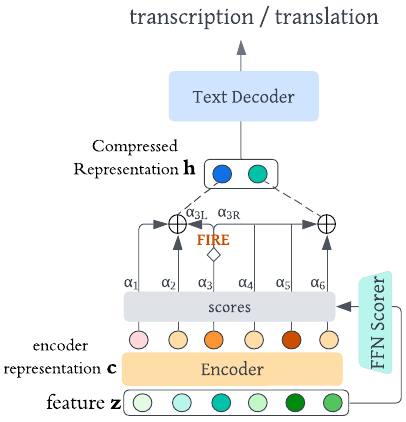}
    \caption{Illustration of Continuous Integrate and Fire.}
    \label{figure::cif_design}
\end{figure}

\noindent Continuous Integrate and Fire \citep[\text{CIF}]{dong2020cif} predicts a score for each position and dynamically aggregates the semantic representation. As shown in \cref{figure::cif_design}, CIF first computes a list of scores $\bm \alpha$ similar to our proposed method. Then, starting from the first position, it accumulates the scores (and representation) until reaching a pre-defined threshold\footnote{~we set $\beta=1$ throughout our experiments, following prior work \cite{dong-etal-2022-learning, Chang_2022}} $\beta$. Once reaching the threshold, it \fire the accumulated representation and starts to accumulate again. As shown in \cref{figure::cif_design}, suppose we originally have representation $\bm z = (z_1, \cdots, z_6)$ with corresponding scores $\bm \alpha = (\alpha_1, \cdots, \alpha_6)$. Suppose we reach the threshold at $t=3$, $\ie \alpha_1 + \alpha_2 + \alpha_3 >= \beta$, then we \fire the representation by taking the weighted average of score and representation $c_1 = \alpha_1 * \bm z_1 + \alpha_2 * \bm z_2 + \alpha_{3L} * \bm z_3$. Here $c_1$ becomes the compressed representation for the region $t = [1,3]$. Note that since $\alpha_1 + \alpha_2 + \alpha_3 >= \beta$, we have residual score $\alpha_{3R} = \alpha_1 + \alpha_2 + \alpha_3 - \beta$, which is left for future accumulation, and we only use $\alpha_{3L} = \alpha_3 - \alpha_{3R}$ when weighting representation $\bm z_3$. More generally, suppose the previous \fire occurs at position $j$ and at current step $i$ the accumulated score reaches the threshold, the aggregated representation is computed as 
\begin{equation}
    h = \alpha_{jR} * \bm z_j + \sum_{t=j+1}^{i-1} \alpha_t * \bm z_t + \alpha_{iL} * \bm z_i
\end{equation}

To enforce the compression rate $r$, we follow \cite{dong-etal-2022-learning, Chang_2022} to re-scale the predicted scores so that the threshold $\beta$ is reached $T_y$ times when accumulating the scores:
\begin{align}
\alpha_t &= \sigma(s_t) \\
\tilde\alpha_t &= \frac{\beta n^*}{\hat{n}} \alpha_t = \frac{\beta \cdot T_y}{\sum_{t=1}^{T_x}\alpha_t} \alpha_t \label{eq::rescale}
\end{align}
Here $\sigma$ is the sigmoid function and $\hat{n}$ is the normalization term (summation of un-scaled scores) and $n^*$ denotes the number of desired selections, \ie $n^* = T_y$. We assume the input feature is longer than the output ($T_x > T_y$), so re-scaling scores to \writeaction $T_y$ means we employ a dynamic compression rate $r = T_x / T_y$ while transducing the streams. Note that $T_y$ is only observed during training and we cannot re-scale $\bm s$ in test time. Therefore, we adopt a length penalty loss \cite{Chang_2022, dong-etal-2022-learning} during training to regularize the segmenter to ensure proper learning of segmentations:
\begin{equation}
    \mathcal{L}_{\text{lp}}(\bm X, \bm Y;\theta) = (n^* - \hat{n})^2 = \left(T_y - \sum_{t=1}^{T_x} \sigma\left(F_{\rm seg}(\bm x_t)\right) \right)^2
    \label{eq::lp_loss}
\end{equation}
Finally, our training objective is the combination of negative log-likelihood and length penalty loss:
\begin{equation}
    \mathcal{L}(\bm X, \bm Y; \theta)  = \mathcal{L}_{\text{NLL}}(\bm X, \bm Y;\theta) + \gamma \mathcal{L}_{\text{lp}}(\bm X, \bm Y;\theta)
\end{equation}\label{eq::all}
In practice, the segmenter is only trained for a few thousand steps (so is the length penalty loss) and we set $\gamma = 0.01$.

Our method is fundamentally different because of how we treat the scorer and how we perform compression. In CIF, the compression is performed as an aggregation (weighted average) within each segmented block (decided by the scores and threshold). In \modelname, we directly take out representations and we force the semantic encoder to condense information to those important positions. In other words, \textbf{we did not explicitly perform aggregation like CIF but expect the semantic encoder to learn such aggregation innately through training}. 

Another key difference is how the scorer is learned. In CIF, the weighted average with scores and representation allows a gradient to flow through the scorer. For \modelname, we inject the scores into cross-attention to update the scorer. The major advantage of our approach is that the importance of position is \textbf{judged by the attention from the decoder to the encoder representation}, which helps segment the speech representation in the way that the text decoder perceives it.

For more details, we direct readers to the prior work \cite{dong2020cif, dong-etal-2022-learning, Chang_2022}.

\begin{table*}[ht]
\resizebox{\linewidth}{!}{
\begin{tabular}{l|ccccccccccc}
    \toprule
    \multirow{2}{*}{\textbf{Model}} & \multicolumn{11}{c}{\textbf{Noise Ratio}} \\
    \cmidrule{2-12}
    & 0\% & 5\% & 10\% & 15\% & 20\% & 25\% & 30\% &35\% &40\% &45\% &50\%\\ 
    \midrule
    Vanilla S2T &  \textbf{15.0} & \textbf{18.7} & \textbf{23.6} & \textbf{29.6} & \textbf{34.0} & \textbf{38.3} &  43.7 & 46.8 &  51.8 & 56.1 & 61.0 \\ 
    \midrule
    S2T + CNN & \worst 24.2 & \worst 26.7 & \worst 31.9 & \worst 37.1 & \worst 40.8 & \worst 45.5 & \worst 49.9 &\worst 53.6 &\worst 58.9 &\worst 61.6 &\worst 65.3 \\
    S2T + CIF & \worse 16.8 & \worse 20.4 & \worse 26.0 & \worse 30.7 & \worse 36.0 & \worse 40.1 & \worse 44.9 &\worse 50.1 & \worse 53.9 &\worse 58.9 & \worse 62.2 \\ 
    S2T + \modelname & \better 15.9 & \better 19.7 & \better 25.1 & \better 29.8 & \better 34.7 & \better 38.6 & \better \textbf{41.8} & \better\textbf{46.7} &\better\textbf{51.0} &\better\textbf{55.6} &\better\textbf{60.1} \\
    \bottomrule
\end{tabular}}
\caption{Word Error Rate of models given the noise injection ratio from 0\% to 50\%. Best numbers are \textbf{bolded} and better results are highlighted by the \textcolor{myBlue}{blue boxes} while bad results are highlighted in \textcolor{myDeepYellow}{yellow boxes}. Compared to other compression methods, our proposed \modelname ~is the most robust model across all noise injection ratios. When the noise ratio reaches beyond 30\%, \modelname ~even outperforms the S2T model without compression. All compression models are trained with the compression rate 12.}
\label{table::robustness}
\vspace{-1em}
\end{table*}

\section{Noise Injection}\label{app::noise_injection}

In this section, we test the robustness of compression methods when noise is injected into the original clean speech from \libritts. Instead of using synthetic signals such as Gaussian noise, we follow \citet{zeghidour2021soundstream} to use natural noise (\eg noise from the air conditioner, shutting door, etc.,) from Freesound\footnote{~We download the audio file for different noise from \url{https://github.com/microsoft/MS-SNSD}} \cite{freesound}. We vary the ratio of noise injection from 5\% to 30\%, as shown in \cref{table::robustness}. Given a ratio, we first calculate the duration of noise $L$ (\eg if the ratio is 0.1 and speech is 10 seconds, then we inject $L=1$ second of noise) and randomly select a range of length $L$ from the clean speech to inject noise. 
As shown in \cref{table::robustness}, as the noise ratio increases, \modelname ~has the smallest degradation and consistently outperforms CIF and CNNs. After reaching noise ratio $\geq 30\%$, \modelname ~even outperforms the vanilla S2T model without compression. Such findings show that \modelname ~has a more robust performance with the help of anchor representation, making it suffer less from noise injection and obtain better ASR performance.

\section{Similarity Test with Compressed Representation}\label{app::sts}

In \cref{sec::compress_asr_result}, we show \modelname's superior performance on ASR, demonstrating the effectiveness of condensing information to a few positions for the text decoder. In this section, \textbf{we evaluate speech representation's similarity to further probe the quality of the compressed representation, without being influenced by the decoder}. More specifically, we use the test set of LibriTTS and for each English transcription, we compute its cosine similarity score against all other transcriptions, using a pre-trained sentence-transformer encoder\footnote{~In practice, we use public checkpoint from: \url{https://huggingface.co/sentence-transformers/all-MiniLM-L6-v2}} (it computes a sentence-level representation from BERT and perform mean pooling to obtain a uni-vector representation). We regard the ranking from sentence-Transformer's similarity as ground truth (as the transcriptions are non-complex English sentences); then we use our speech semantic encoders to compute cosine similarity for all pairs of speech representations and verify if the ranking is similar to the ground truth. 

For the baseline vanilla S2T model, we perform mean pooling (MP) on its encoder representation $\bm c$ to obtain a uni-vector representation for each speech input and compute cosine similarities. For the other three models with compression, we first obtain the compressed representation $\bm h$ and we try two approaches to compute similarity. The first approach is the same as the baseline, where we apply MP on the compressed representation to obtain uni-vector representations. The second approach is inspired by the MaxSim (MS) algorithm used in ColBERT \cite{Khattab2020ColBERTEA}, which computes the average of maximum similarity across the compressed representations.

Then we measure the quality of our trained speech semantic encoders with metrics widely used in retrieval and ranking--Normalized Discounted Cumulative Gain (nDCG) and Mean Reciprocal Rank (MRR). From the results shown in \cref{table::speech_similarity}, \modelname ~still obtains the best-performing representation, with $\text{MRR}@10 = 0.087, \text{nDCG}@10=0.453$. Note that the performance is not very high as we did not train the model specifically for the sentence similarity task. Rather, we used the similarity task as an intrinsic measurement for the quality of condensed representations \textbf{to exclude the influence of the text decoder}. 

Comparing the numbers in \cref{table::speech_similarity}, \modelname ~consistently obtains better speech representation (for both MP and MS algorithms) for the similarity task. Interestingly we find that \modelname-30's representation works better in mean pooling compared to \modelname-12, suggesting that more condensed information works better for mean pooling. However, the MaxSim algorithm better leverages the multi-vector representation, which enables \modelname-12 to obtain the best ranking performance.

    

\begin{table}[h]
    \centering
    \begin{tabular}{lcccc}
    \toprule
    & \multicolumn{2}{c}{\textbf{NDCG @ 10}} & \multicolumn{2}{c}{\textbf{MRR @ 10}} \\
    \cmidrule(lr){2-3} \cmidrule(lr){4-5}
    \textbf{Model} & \textbf{MP} & \textbf{MS} & \textbf{MP} & \textbf{MS} \\
    \midrule
    Vanilla S2T & 0.407 & N/A & 0.053 & N/A \\ \midrule
    Conv-12 & 0.399 & 0.41 & 0.035 & 0.053 \\
    CIF-12 & 0.418 & 0.444 & 0.056 & 0.078 \\ 
    \modelname-12 & 0.429 & \textbf{0.453} & 0.064 & \textbf{0.087} \\ \midrule
    \modelname-18 & 0.429 & 0.446 & 0.055 & 0.078 \\
    \modelname-30 & \textbf{0.437} & 0.441 & \textbf{0.078} & 0.08 \\
    \bottomrule
    \end{tabular}
    \caption{Performance of speech rankings by different representation. \modelname ~achieves the best performance as evaluated by NDCG@10 and MRR@10. The best performance is achieved through the MaxSim algorithm; interestingly, \modelname-30 achieves the best performance with the Mean Pooling algorithm.}
    \label{table::speech_similarity}
\end{table}

\begin{table*}[h]
\centering
\small
\begin{tabular}{@{}lccccccccc@{}}
\toprule
\multirow{2}{*}{\textbf{Stage}} & \multirow{2}{*}{\textbf{Batch Size}} & \multirow{2}{*}{\textbf{Seq Len}} & \multirow{2}{*}{\textbf{No Compression}} & \multicolumn{4}{c}{\textbf{With Compression}} \\ 
\cmidrule(r){5-8} &  &  & & \textbf{r=2} & \textbf{r=5}& \textbf{r=10} & \textbf{r=20} \\ 
\midrule
Inference & 1 & 1000 & 1196 & 1076 & 1004 & 987 & 970 \\ 
& 1 & 2000 & 2975 &2509 & 2237 & 2138 & 2101   \\ 
& 1 & 3000 & 5744 & 4736 & 4101 & 3894 & 3739  \\ 
 & 1  & 4000 & 9540 & 7711 & 6637 & 6269 & 6102          \\
  & 1  & 5000 &  14314 &11493 &9805 &9237 &8951    \\
& 1  & 6000 & OOM  & OOM & 13587 & 12786  & 12434 \\ 
\midrule
Training  & 128 & 100  & 4964 & 4730 & 4209 & 4160& 4124         \\ 
  & 128 & 200  & 10687 & 9948 & 9465 & 9302& 9223         \\
\bottomrule
\end{tabular}
\caption{Memory usage (MB) of the encoder-decoder model with and without our proposed compression method. OOM: out of memory.}
\label{table::memory}
\end{table*}

\section{Memory Usage Benchmark}\label{app::memory}
In this section, we describe our setup to benchmark memory usage, which compares our proposed approach with a vanilla encoder-decoder model that does not support compression. We use Google Colab with a runtime that uses a T4 (16G memory) GPU. Then for each experiment, we run it 5 times and report the average in \cref{table::memory}. Both encoder and decoders follow our setup in \cref{app::hyper}, except that the encoder's maximum position is increased to 8,196 to support the benchmark experiment with long sequences. Note that the sequence length reported is the length of the input feature (which we compress by $r \in \{2, 5, 10, 20\}$). We set the output sequence's length to be $\frac{1}{10}$ of the input, similar to the ratio in our simultaneous \emph{speech-to-text} experiments.

\section{Hyper-parameters}\label{app::hyper}
\begin{figure}[h]
    \centering
    \includegraphics[scale=0.9]{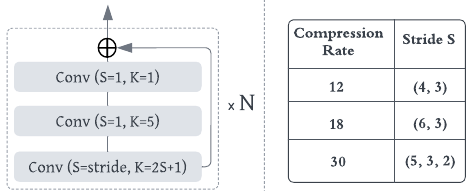}
    \caption{\textit{Left}: Blocks of CNNs used to compress representation. \textit{Right}: Stride sizes we used in experiments for different compression rates.
    }
    \label{figure::conv_design}
\end{figure}

We provide hyper-parameters used for model configuration and training in this section. For different compression rates, the CNNs' stride configuration is shown in \cref{figure::conv_design}. For example, a stride of (4,3) means we stack two CNN blocks, one with stride 4 and another with stride 3, achieving a compression rate of 12.

In this section, we provide the hyper-parameters and training configurations for all our experiments. We use a hidden dimension of 512 across all models. The tokenizer is developed using Byte Pair Encoding \citep[BPE]{bpe}, with a vocabulary size of 10,000. The segmenter is parameterized by a 2-layer FFN with ReLU \cite{Agarap2018DeepLU} activation in between; the first FFN has input and output dimensions both set to 512 and the second FFN has input dimension 512 with output dimension 1. Our experiments are conducted using the Adam optimizer, configured with $\beta_1 = 0.9$ and $\beta_2 = 0.999$. These experiments are conducted with a data-parallel setting with 4 A100 GPUs.

For the audio processing, we set the sampling rate to 16,000. In the encoder configuration, we use a maximum of 1,024 positions for Automatic Speech Recognition (ASR) and 2,048 for Speech Translation (ST), with each encoder consisting of 4 layers and 8 attention heads. The decoder mirrors the encoder in its architecture, with 4 layers and 8 attention heads, but differs in its maximum positions, set at 512, and its vocabulary size, also at 10,000.

For non-streaming ASR in our pre-training setup, both the encoder and decoder are trained to converge with a learning rate of 1e-4, a batch size of 32, and a warmup of 10,000 steps. Subsequently, the compression module (CNN/CIF/\modelname) is fine-tuned using a learning rate of 5e-5 alongside the pre-trained encoder and decoder. The segmenter is trained for 6,000 steps with feedback from the encoder-decoder's cross-attention, as discussed in \cref{sec::method}, after which it is frozen. Post this, we further fine-tune the encoder and decoder until convergence.

For streaming \emph{speech-to-text} tasks, the feature extractor (\wav), encoder, and decoder are jointly trained with a learning rate of 5e-5, a batch size of 8, and gradient accumulation every 4 steps. A causal mask is added to \wav during this process. Following convergence, the compression module undergoes fine-tuning using a learning rate of 5e-5 and a batch size of 16. Similar to the non-streaming setup, the segmenter is updated only in the first 6,000 steps.

\section{Differentiable Average Lagging}\label{app::dal}
Consider a raw speech with length $T_x$ which is segmented into $|X|$ chunks. We define the length of $i^{th}$ segment (chunk) as $|X_i|$ (so that $|X| = \sum_{j=1}^{|X|} |X_j|$), and we define $d_i = \sum_{t=1}^i |X_t|$ as the total time that has elapsed until $i^{th}$ speech segment $X_i$ is processed. With the aforementioned notation, DAL is defined to be:
\begin{equation}\label{eq::dal}
    \text{DAL} = \frac{1}{T_y} \sum_{i=1}^{T_y} d_i' - \frac{i-1}{\gamma}
\end{equation}
where $T_y$ is the length of text tokens and  $1/\gamma$ is the minimum delay after each operation, computed as $1/\gamma = \sum_{j=1}^{|X|} |X_j|/T_y$ (\ie the averaged elapsed time for each token is used as the minimum delay). Lastly, $d_i'$ is defined as:
\begin{equation}
    d_i' = \begin{cases}
        d_i & i=0\\
        \text{max}(d_i, d_{i-1}' + 1/\gamma) & i > 0
    \end{cases}
\end{equation}

The smaller the DAL, the better the system in terms of latency. For more discussions for DAL and latency-quality trade-off in SimulST, we direct readers to prior work \cite{ma-etal-2020-simuleval, milk} for more details.

\end{document}